\documentclass[english,11pt]{article}

\usepackage{amsthm}
\usepackage{amsmath}
\usepackage{natbib}
\usepackage{fullpage}
\usepackage[hidelinks]{hyperref}
\usepackage{graphicx}
\providecommand{\keywords}[1]{\noindent \textbf{Keywords} #1}
\usepackage[normalem]{ulem}
\usepackage{multirow}
\usepackage[shortlabels]{enumitem}
\usepackage[plain,noend]{algorithm2e}
\usepackage{mathtools,xparse}
\usepackage{amsfonts}

\def\EE{{\mathbb{E}}}
\def\RR{{\mathbb R}}
\def\NN{{\mathbb{N}}}

\def\wt{\widetilde}

\newcommand{\C}{\mathcal{C}}

\newcommand{\N}{\mathrm{Normal}} 

\renewcommand{\S}{\mathcal{S}}

\newcommand{\U}{\mathrm{Uniform}}

\newcommand{\Z}{\mathcal{Z}}

\def\mcmc{\mathrm{mcmc}}

\renewcommand{\NN}{\mathbb{N}}
\newcommand{\OO}{\mathcal{O}}
\renewcommand{\RR}{\mathbb{R}}

\newcommand{\IG}{\mathrm{IG}}

\newcommand{\iid}{\stackrel{\rm iid}{\sim}}

 \DeclareMathOperator*{\argmin}{\arg\!\min}
 
\def\1{\mathbf{1}}
 
\def\scaleinvchi2{\mathrm{Scale}\text{-}\mathrm{inv}\text{-}\chi^2}

\def\I{\mathbb{I}}

\def\b0{\mathbf{0}}

\usepackage{cleveref}
\crefname{assumption}{Assumption}{Assumptions}
\crefname{remark}{Remark}{Remarks}
\crefname{proposition}{Proposition}{Propositions}
\crefname{corollary}{Corollary}{Corollary}
\crefname{theorem}{Theorem}{Theorems}
\crefname{section}{Section}{Section}
\crefname{lemma}{Lemma}{Lemma}
\crefname{table}{Table}{Table}
\crefname{algorithm}{Algorithm}{Algorithms}
\crefname{example}{Example}{Examples}
\crefname{figure}{Figure}{Figure}
\crefname{appendix}{Appendix}{Appendix}
\crefname{equation}{equation}{equation}

\newtheorem{theorem}{Theorem}
\newtheorem{corollary}{Corollary}
\newtheorem{proposition}[theorem]{Proposition}

\def\log{{\rm log}}

\newcommand{\blue}[1]{{#1}}
\def\bY{\mathbf{Y}}
\def\bX{\mathbf{X}}

\def\ahat{\widehat{a}}

\def\bpi{\pi}

\def\bq{\mathsf{q}}
\def\obq{\overline{\bq}}
\def\obpi{\overline{\bpi}}
\def\onabla{\overline{\nabla}}

\def\buffer{B}
\def\Xdomain{\Omega_X}
\def\Xdomrare{\Omega_{\mathrm{rare}}}

\def\hatnabla{\widehat{\nabla}}

\def\mb{\S}

\def\var{\mathrm{var}}

\def\latidx{k}

\def\trans{A}

\def\wt{\widetilde}

\def\atilde{\widetilde{a}}
\def\ybar{\overline{Y}}
\def\MAP{\mathrm{MAP}}
\def\mcmc{\mathrm{MCMC}}
\graphicspath{{Figures/}}
\usepackage{amsmath,amssymb,subcaption,natbib,algorithmic,times, comment,mathrsfs,placeins,bbm,float,hyperref,xcolor,amsthm,inputenc,rotating,hyperref}

\title{Targeted stochastic gradient MCMC for HMMs with rare latent states}
\author{
Rihui Ou$^{1}$ 
\and 
Alexander Young$^{2}$\footnote{Corresponding author; {alexander\_young@fas.harvard.edu}.}
\and 
Deborshee Sen$^{3}$
\and 
David Dunson$^1$ 
}
\date{$^1$Department of Statistics, Harvard University 
\\
$^2$Department of Statistical Science, Duke University
\\
$^3$Google}

\begin{document}
\maketitle

\begin{abstract}
\noindent 
Markov chain Monte Carlo (MCMC) algorithms for hidden Markov models often rely on the forward-backward sampler. This makes them computationally slow as the length of the time series increases, motivating the  development of sub-sampling-based approaches. These approximate the full posterior by using small random subsequences of the data at each MCMC iteration within stochastic gradient MCMC. In the presence of imbalanced data resulting from rare latent states, subsequences often exclude rare latent state data, leading to inaccurate inference and prediction/detection of rare events. We propose a targeted sub-sampling (TASS) approach that over-samples observations corresponding to rare latent states when calculating the stochastic gradient of parameters associated with them. TASS uses an initial clustering of the data to construct subsequence weights that reduce the variance in gradient estimation. This leads to improved sampling efficiency, in particular in settings where the rare latent states correspond to extreme observations. We demonstrate substantial gains in predictive and inferential accuracy on real and synthetic examples. 
\end{abstract}

\section{Introduction} \label{sec.inntro}

Hidden Markov models (HMMs) are widely used due to their versatility in characterizing a breadth of phenomena; this includes 
credit ratings \citep{petropoulos2016novel},
protein folding dynamics \citep{peyravi2019composite,protein_hmm}, 
speech recognition \citep{muhammad2018speech,speech_recog_2}, 
stock market forecasting \citep{zhang2019enhancing}, 
and rare event detection \citep{galagedarage2019process,damicis2023}. 
Computation for HMMs often relies on the forward-backward algorithm \citep{baum1972inequality}. This includes Monte Carlo \citep{scott2002bayesian} and variational methods \citep{johnson2014stochastic} among Bayesian approaches, and expectation-maximization \citep{bishop2006pattern} among frequentist approaches. At each iteration, a local update of the unknown latent states is performed using forward-backward to obtain their marginal distributions. This is followed by a global update for the parameters of the emission distributions. The forward-backward algorithm passes through the entire sequence of observations, resulting in a computational complexity that is at least linear in the number of observations. 
However, even linear scaling can be computationally prohibitive for very long data sequences.

Obtaining reliable uncertainty quantification is particularly important in scientific applications. Along with the increasing collection of very long time series data, this has inspired the development of scalable Bayesian approaches for inference on hidden Markov models, including techniques based on sub-sampling \citep{johnson2014bayesian, hughes2015scalable} and data-thinning \citep{hunt2018online}. 
Sub-sampling relies on using a small subsequence of the observations at each iteration of an algorithm. The past decade has witnessed the development of a variety of posterior sampling algorithms based on data sub-sampling for independent observations \citep{welling2011bayesian, chen2014stochastic, bouchard2018bouncy, bierkens2019zig, quiroz2019speeding,poisson_subsamp}. 
Particularly relevant to this article are methods that use non-uniform data sub-sampling in combination with MCMC. Examples include coresets \citep{huggins2016coresets, campbell2019automated, fast_coresets}, minibatches with importance sampling \citep{csiba2018importance,li2020improving},
stochastic gradient MCMC \citep{needell2014,kmeansSGMCMC,dalalyan2019user,sgmcmc_review}, and non-reversible piecewise deterministic Markov processes
\citep{sen2020efficient}. Also relevant is work using non-uniform sub-sampling in stochastic gradient optimization \citep{needell2014, johnson2018training}. 

However, the sequential nature of hidden Markov models brings in additional complications. In relevant literature, \cite{ma2017stochastic} developed a stochastic gradient Markov chain Monte Carlo (MCMC) framework based on stochastic gradient Langevin dynamics (SGLD; \citealp{welling2011bayesian}), and \cite{foti2014stochastic} developed a stochastic variational inference algorithm. These algorithms exploited the underlying mixing of the latent states to motivate sub-sampling based on short \blue{and medium length} time blocks, \blue{which has been a common approach in recent work on variational Bayes \citep{jung2022,chen2023}, divide-and-conquer \citep{wang2023}, and optimization-only \citep{sidrow2024} approaches to HMMs}. \cite{salomone2020spectral} developed a sub-sampling MCMC algorithm for a class of stationary time series models, but without considering HMMs. 


We consider discrete state-space hidden Markov models in this article, motivated in particular by the scenario in which there are rare latent states. Indeed, a typical focus of hidden Markov model analyses is inferences on and prediction of rare events, which are observations corresponding to rare latent states. 
There are diverse examples of rare latent states in the literature ranging from network attacks \citep{ourston2003applications} to solar flares \citep{hall2015online} to stuck pipes in oil drilling operations \citep{damicis2023}. Competing algorithms for scalable posterior sampling in HMMs, such as in \cite{ma2017stochastic}, will commonly fail to include any observations from the rare states in the sampled subsequences. This leads to very high posterior uncertainty in estimating parameters related to the rare latent states and commonly failure to identify the presence of a rare latent state entirely (\cref{fig:paramest_2rare}).

Motivated by these issues, we propose a targeted sub-sampling approach for stochastic gradient MCMC for hidden Markov models (TASS-HMM). Instead of using a single subsequence for all components of the stochastic gradient estimate, we instead consider different subsequences for parameters corresponding to rare latent states and those corresponding to common states. Our approach is ultimately based on clustering of the observations to over-sample subsequences containing rare latent states for the former estimates, and then adjusting for biased sub-sampling in conducting stochastic gradient MCMC. 
TASS drastically improves the accuracy of gradient estimates, which is known to lead to improved performance for stochastic gradient MCMC \citep{dalalyan2019user}. 
We use synthetic data to numerically demonstrate the advantages of TASS-HMM. We also apply TASS-HMM to analyze solar flare \blue{and sleep cycle data}, which have rare spikes.
While we focus on Langevin dynamics in this article, the TASS approach can also be used with Hamiltonian dynamics \citep{chen2014stochastic, dang2019}. 

The rest of the article is organised as follows. 
We introduce the necessary background in \cref{sec.background}. We present the proposed targeted sub-sampling scheme in \cref{sec.c.sg.mcmc}. 
\cref{sec.numerics} is dedicated to numerical experiments on synthetic data. In particular, we consider situations with one, multiple, and no rare latent states (Sections \ref{sec.single.rare}, \ref{sec.multiple.rare}, and \ref{sec.norare.state} respectively), a run-time analysis (\cref{sec.run.time.analysis}), comparison with full Langevin dynamics MCMC (\cref{sec.comparison.full.mcmc}), and a gradient estimation analysis (\cref{sec.grad.est}).
\cref{sec.real.data} contains two real data analyses: inference from solar flare data in \cref{sec.solar.flare} and from sleep cycle data in \cref{sec.sleep.cycle}. Finally, \cref{sec.discussion} concludes. 

\section{Background} \label{sec.background} 

\subsection{Hidden Markov models}
\label{sec.hmm.background}

A hidden Markov model (HMM) consists of latent states $\{X_t\}_{t=0}^T \in \Xdomain$ which form a Markov chain, and corresponding observations $\{Y_t\}_{t=1}^T$. In this article, we consider a discrete state-space, that is, $\Xdomain = \{1,\dots,K\}$. We assume that $A = ((A_{kk'})_{k=1}^K)_{k'=1}^K$ is the transition matrix of the latent states, that is, $A_{kk'} = P(X_t=k\mid X_{t-1} = k')$ for $k,k' = 1, \dots K$. We also suppose that $Y_t \mid (X_t = k)$ is parameterized by $\phi_k$ for $k=1,\dots,K$, with $\phi = \{\phi_k\}_{k=1}^K$ being the emission parameters.
The joint distribution of $\bY = \{Y_t\}_{t=1}^T$ and $\bX = \{X_t\}_{t=0}^T$ factorizes as $p(\bX,\bY\mid A,\phi ) = p(X_0) \prod_{t=1}^T p(Y_t\mid X_t,\phi) p(X_t\mid X_{t-1},A)$, where $p(X_0)$ is the distribution of the initial state $X_0$. 

Throughout this article, we assume that the latent Markov chain is recurrent and irreducible, and that the latent states are at stationarity, so that $X_0$ is drawn from the unique stationary distribution for $\trans$, which is a common assumption in HMM inference \citep{ma2017stochastic}. We let $\theta = \{A, \phi_1, \dots, \phi_K\}$ denote all the parameters of the model. For notational convenience, we shall flatten $\theta$ and view it as a vector in $\RR^d$, where $d$ is the sum of the dimensions of $A$ and $\phi_1, \dots, \phi_K$. We shall use the notation $p_\theta(Y_t \mid X_t)$ and $p_\theta(X_t \mid X_{t-1})$ for $p(Y_t\mid X_t,\phi)$ and $p(X_t\mid X_{t-1},A)$, respectively. 

\subsection{Stochastic gradient MCMC for hidden Markov models}
\label{sec.sg.mcmc}

The latent states $\bX$ can be marginalized out to obtain likelihood $$p(\bY\mid \theta) = \1^\top \left\{ \prod_{t=1}^T \blue{P_\phi}(Y_t) A \right\} \pi,$$ where $\blue{P_\phi}(Y_t)$ is a diagonal matrix with entries $\left(P_{\phi}(Y_t)\right)_{kk} = p(Y_t\mid X_t = \latidx,\phi)$, $\latidx=1,\dots,K$, $\1$ is a $K$-dimensional vector of ones, and $\pi\in\Delta^K$ is the stationary distribution of $X_t$, with $\Delta^K$ the probability simplex.
Given a prior $p_0(\theta)$ for $\theta$, the posterior of $\theta$ given observations $\bY$ is 
\begin{equation}
p(\theta\mid\bY) = \frac{p(\bY \mid \theta) p_0(\theta)}{p(\bY)} \propto p(\bY\mid\theta) p_0(\theta).
\label{eq.posterior}
\end{equation}
The posterior can be written in the form $p(\theta \mid \bY) \propto \exp\{- U(\theta)\}$, where $$U(\theta) = \\ -\log \left[\1^\top \left\{ \prod_{t=1}^T \blue{P_\phi}(Y_t)A\right\} \pi\right] - \log\, p_0(\theta)$$ 
is known as the potential function. 
Assuming a continuous parameter $\theta$, marginalizing out over the latent states allows one to use gradient-based sampling algorithms for posterior inference; these include Hamiltonian Monte Carlo \citep{duane1987hybrid} and Langevin-based algorithms \citep{roberts1996exponential,welling2011bayesian}. Using the notation $\nabla_i$ for $\partial/(\partial \theta_i)$, these require calculating the gradient of $U(\theta)$, which can be written as 
\begin{equation} \label{eq.full.grad}
\nabla_i U(\theta) 
= 
-\frac{\sum_{t=1}^T \bq_{t+1}^\top \left \{ \nabla_i \blue{P_\phi}(Y_t) A \right \} \bpi_{T-1}}{\1^\top \{ \prod_{t=1}^T \blue{P_\phi}(Y_t)A \} \pi}
-
\nabla_i \log p_0(\theta),
\end{equation}
where 
\begin{equation} \label{eq.true.buffer}
\bq_{t+1}^\top 
= 
\1^\top \prod_{s=t+1}^T \{ \blue{P_\phi}(Y_s)A \}
~~ \text{and} ~~ 
\bpi_{T-1} 
=
\left \{ \prod_{s=1}^{T-1} \blue{P_\phi}(Y_s)A \right \} \pi.
\end{equation}

Evaluating the gradient \eqref{eq.full.grad} becomes increasingly computationally expensive for large $T$ due to the repeated matrix multiplications needed to calculate $\bq_{t+1}^\top$ and $\bpi_{T-1}$ in \cref{eq.true.buffer}. To address this issue, \cite{ma2017stochastic} utilize subsequences to construct gradient approximations. We describe a slightly simplified version of their approach in the remainder of this section. 

Consider a half-width $L \ll T$ of the subsequences and fix subsequence centers $\C = \{L+1, 3L+1, \dots, T-L\}$. For $\tau \in \C$, consider the subsequences $\bY_\tau = (Y_{\tau - L}, \dots, Y_\tau,\dots, Y_{\tau+L})$ of length $(2L+1)$ centered at $t=\tau$, and let $\blue{P_\phi}(\bY_\tau ) = \prod_{t=\tau-L}^{\tau + L} \{\blue{P_\phi}(Y_t)A\}$. The full chain $Y_{1:T}$ is thus partitioned into $N=T/(2L+1)$ sequential subsequences such that $\bY = \{ \bY_{\tau_1},\dots, \bY_{\tau_N}\}$.
Equation \eqref{eq.full.grad} can then be expressed using these subsequences as
\begin{equation} \label{eq.grad.sub.series}
\nabla_i U(\theta) 
= 
-\sum_{n=1}^{N} \frac{\bq_{\tau_n + L+1}^\top \left \{\nabla_i \blue{P_\phi}(\bY_{\tau_n}) \right \} \bpi_{\tau_n -L-1} }{\bq_{\tau_n +L+1}^\top \, \blue{P_\phi}(\bY_{\tau_n}) \, \bpi_{\tau_n -L-1}}
-
\nabla_i \log \, p_0(\theta).
\end{equation}
This suggests constructing a stochastic gradient estimate using only one term of the sum in the right hand side of \cref{eq.grad.sub.series}. 
Unfortunately, calculating $\bq_{\tau_n +L+1}^\top$ and $\bpi_{\tau_n -L-1}$ are still computationally expensive when $T$ is large as they must pass over the full series after and before $\bY_{\tau_n}$, respectively, thereby requiring further approximations.

Suppose that the transition matrix $\trans$ is known, and let $\buffer$ be a buffer length longer than the inverse of the spectral gap of $\trans$. Observations in $\bY_\tau$ and those more than $\buffer$ steps before or after $\bY_\tau$ will be approximately independent, leading to the approximations $\bq_{\tau_n +L+1}^\top \approx \obq_{\tau_n +L+1}^\top = \1^\top [ \prod_{s = \tau_n +L+1}^{\tau_n +L+1+\buffer} \{ \blue{P_\phi}(Y_s) A \}]$ and $\bpi_{\tau_n -L-1} \approx \obpi_{\tau_n -L-1} = [ \prod_{s = \tau_n -L-1-\buffer}^{\tau_n -L-1} \{ \blue{P_\phi}(Y_s) A \} ] \pi$; these are truncations of the products in \cref{eq.true.buffer}, making them computationally cheaper to calculate.
Using these approximations, we define 
\begin{align} \label{eq.grad.sub.series.bar}
\onabla_i U(\theta)
& =
- \sum_{n=1}^{N} \frac{\obq_{\tau_n + L+1}^\top \left \{\nabla_i \blue{P_\phi}(\bY_{\tau_n}) \right \} \obpi_{\tau_n -L-1} }{\obq_{\tau_n +L+1}^\top \, \blue{P_\phi}(\bY_{\tau_n}) \, \obpi_{\tau_n -L-1}}
-
\nabla_i \log \, p_0(\theta).
\end{align}
In practice, methods to approximate the spectral gap of the unknown stochastic matrix $A$ are given in \cite{ma2017stochastic}.

As alluded to earlier, a random sub-sample of $S$ subsequences is also used in the calculation of \cref{eq.grad.sub.series}. Choosing a subsequence $\S = \{J_1, \dots, J_M\} \subset \{1, \dots, N\}$ of size $M$ such that $J_1,\dots,J_M \iid \U(\{1,\dots,N\})$ leads to an approximated gradient $\hatnabla U(\theta)$ with components
\begin{equation} \label{eq.grad.approx.mb}
\hatnabla_i U(\theta) 
=
- \frac{N}{M} \sum_{m=1}^M \frac{\obq_{\tau_{J_m} +L+1}^\top \left \{\nabla_i \blue{P_\phi}(\bY_{\tau_m}) \right \} \obpi_{\tau_{J_m} -L-1} }{\obq_{\tau_{J_m} +L+1}^\top \, \blue{P_\phi}(\bY_{\tau_{J_m}}) \, \obpi_{\tau_{J_m} -L-1}}
-
\nabla_i\log \,p_0(\theta);
\end{equation} 
we refer to this as the uniform sub-sampling strategy in the sequel.

With these gradient approximations in hand, we summarize \cite{ma2017stochastic}'s stochastic gradient MCMC algorithm for hidden Markov models in \cref{alg.sg.mcmc.hmm}; this is based on the stochastic gradient Langevin dynamics (SGLD) algorithm of \cite{welling2011bayesian}. 

\begin{algorithm}
\caption{Stochastic gradient MCMC for hidden Markov models \citep{ma2017stochastic}.}
\label{alg.sg.mcmc.hmm}

\textbf{Input:} Observations $Y_1,\dots,Y_T$, buffer length $B$, half-width $L$, step-size $\epsilon$, initial value $\theta_0$, number of posterior samples $Z$, subsequence size $M$.

\begin{algorithmic}[1]

\FOR{z = 1 \TO Z}

\STATE 
Draw subsequence $\mb$ of size $M$.

\STATE 
Compute gradient approximation $\hatnabla U(\theta_{z-1})$ using \cref{eq.grad.approx.mb}.

\STATE 
Set $\theta_z = \theta_{z-1} - (\epsilon/2) \hatnabla U(\theta_{z-1}) + \xi_z$ for $\xi_z \sim \N(0,\epsilon I)$.

\ENDFOR
\end{algorithmic}
\textbf{Output:} Samples $\{\theta_z\}_{z=1}^Z$. 
\end{algorithm}

\subsection{Rare latent states}

If the latent Markov chain has some rare latent states, say $\Xdomrare \subset \{1,\dots,K\}$, the latent chain $\bX$ spends only a small portion of time in $\Xdomrare$. Uniform sub-sampling strategies do not take this issue into account, causing observations $Y_t$ corresponding to $X_t \in \Xdomrare$ to be rare or absent from the sampled subsequences.
As a result, the components of a stochastic gradient estimate associated with parameters pertaining to rare states, such as emission parameters and transition probabilities, will generally contain little to no information from the data. In extreme cases, posterior estimates of these quantities are essentially determined by their priors.
In this article, we consider a clustering-based strategy for deriving targeted non-uniform weights for sampling the subsequences. Our weights are constructed to minimize variance in the stochastic gradient estimates and as such over-sample subsequences with rare states.

\section{Targeted sub-sampling} \label{sec.c.sg.mcmc}

\subsection{Importance sampling in stochastic gradient MCMC} \label{sec.method}

We enumerate the time indices of the set of subsequences as $\{C_1, \dots, C_N\}$, that is, $C_n = \{nL-L+1, \dots, nL+L+1\}$, $n=1,\dots,N$.
Consider weights $a_1,\dots,a_N$ such that $a_n > 0$ and $\sum_{n=1}^N a_n = 1$. An unbiased estimator of the gradient \eqref{eq.grad.sub.series.bar} is given by 
\begin{equation*} 
\hatnabla_i U(\theta) 
=
-\frac1M \sum_{m=1}^M a_{J_m}^{-1} \frac{\obq_{\tau_{J_m} +L+1}^\top \left \{\nabla_i \blue{P_\phi}(\bY_{\tau_m}) \right \} \obpi_{\tau_{J_m} -L-1} }{\obq_{\tau_{J_m} +L+1}^\top \, \blue{P_\phi}(\bY_{\tau_{J_m}}) \, \obpi_{\tau_{J_m} -L-1}}
-
\nabla_i\log \,p_0(\theta)
\end{equation*} 
for $J_1,\dots,J_M \iid \text{Multinomial}\{N, (a_1,\dots,a_N)\}$.

Our goal in this article is to choose the importance weights in an efficient manner.
It is well-understood that reducing the variance of the stochastic gradient $\hatnabla \, \ell(\theta)$ improves the performance of the resulting stochastic gradient sampling algorithm. This has been discussed thoroughly in the context of log-concave posterior distributions \citep{chatterji2018theory,zou2018subsampled,dalalyan2019user}.
In particular, \cite{dalalyan2019user} provides a finite sample bound, which depends linearly on the variance of the gradient estimate, for the Wasserstein-2 distance between the chain and the specified target distribution.
Unfortunately, for HMMs, it is not reasonable to expect the posterior given by \cref{eq.posterior} to satisfy the log-concavity assumption due to the invariance of \cref{eq.posterior} under label switching. 
Other results relax the log-concave assumption but still demonstrate the importance of reducing variance in the stochastic gradient estimates. See for example \cite{raginsky2017non} and \cite{zou2021faster} which provide bounds in Wasserstein-2 and total variation distance respectively, although the initial assumptions therein occlude clear algebraic relationships between convergence rates and the SGLD variance. Nonetheless, these articles collectively provide strong evidence that one should seek to minimize the variance of stochastic gradient estimates when designing an SGLD algorithm.

With this observation in mind, we establish a targeted sub-sampling (TASS) scheme that greatly reduces the variance in the stochastic gradient estimates without adding prohibitive computational costs. 
We fix the number of subsequences $M=1$ in order to simplify the exposition; the approach described can straightforwardly be generalized to $M>1$. 
The contribution of the prior to the potential function $U$ does not depend on the random subsequences, and we can thus ignore it in the analysis.

To begin with, suppose that the latent states are known and are denoted by $x_0, \dots, x_T$. In this case, the log-likelihood is 
$\ell(\theta) = \log \, p_\theta(Y_{1:T},x_{1:T}) = 
\sum_{n=1}^N \sum_{t\in C_n} \{ \log \, p_\theta (Y_t \mid x_t) + \log \, p_\theta(x_t \mid x_{t-1}) \}$,
which has gradient 
$\nabla_\theta \, \ell(\theta) = \sum_{n=1}^N \sum_{t \in C_n} \nabla_\theta \{ \log \, p(Y_t \mid x_t) + \log \, p_\theta (x_t \mid x_{t-1}) \}$.
An unbiased estimator is constructed as $$\hatnabla_\theta \, \ell(\theta) = a_J^{-1} \sum_{t \in C_J} \nabla_\theta \{ \log \, p_\theta(Y_t \mid x_t) + \log \, p_\theta (x_t \mid x_{t-1}) \}$$ for $J \sim \text{Multinomial} \{N, (a_1,\dots,a_N)\}$. 
By \cite{dalalyan2019user}, the importance weights should be chosen to minimize $\EE \| \hatnabla_\theta \, \ell(\theta) - \nabla_\theta \, \ell(\theta)\|^2$, where the expectation is with respect to the randomness in choosing the sub-sample.
%


The following Proposition 1 provides optimal importance sampling weights for this situation. 

\begin{proposition}[Optimal weights for known latent states]
\label{prop.imp.known.latent}
The optimal importance weights when the latent states are known are given by
\begin{equation} \label{eq.optim.weight.known.latent}
\ahat_n 
\propto 
\left ( \sum_{i=1}^d \left [ \sum_{t \in C_n} \nabla_i \left \{ \log \, p_\theta(Y_t \mid x_t) + \log \, p_\theta (x_t \mid x_{t-1}) \right \} \right ]^2 \right )^{1/2}
\end{equation}
for $n=1,\dots,N$, where $\theta = (\theta_1,\dots,\theta_d)$.
\end{proposition}

\begin{proof}[Proof of \cref{prop.imp.known.latent}]
By Condition N of \cite{dalalyan2019user}, the performance of stochastic-gradient MCMC is optimized when 
\begin{align*}
\EE \| \hatnabla_\theta \, \ell(\theta) - \nabla_\theta \, \ell(\theta)\|^2 
& = 
\sum_{i=1}^d \EE \{ \hatnabla_i \ell(\theta) - \nabla_i \, \ell(\theta)\}^2
= 
\sum_{i=1}^d \left [ \EE \{ \{ \hatnabla_i \ell(\theta)\}^2\} - \{\nabla_i \, \ell(\theta)\}^2 \right ]
\end{align*}
is minimized, where the last equality is using $\EE \{ \hatnabla \ell(\theta) \} = \nabla \, \ell(\theta)$. This is equivalent to minimizing $\sum_{i=1}^d \EE [ \{\hatnabla_i \ell(\theta)\}^2]$.
Let $q_\theta(Y_t, x_t, x_{t-1}) = \log \, p_\theta(Y_t \mid x_t) + \log \, p_\theta (x_t \mid x_{t-1})$. Then 
\begin{align*}
& \quad 
\sum_{i=1}^d \EE [ \{\hatnabla_i \ell(\theta)\}^2]
= 
\sum_{i=1}^d \sum_{j=1}^N a_j \left \{ a_j^{-1} \sum_{t \in C_j} \nabla_i \, q_\theta(Y_t, x_t, x_{t-1})\right \}^2
=
\sum_{i=1}^d \sum_{j=1}^N \frac{\gamma_{ij}^2(\theta)}{a_j}
\end{align*}
for $\gamma_{ij}(\theta) = \sum_{t \in C_j} \nabla_i \, q_\theta(Y_t, x_t, x_{t-1})$.
Therefore the optimal importance weights are such that 
\begin{equation*}
\ahat_1, \dots, \ahat_N
=
\argmin_{\substack{a_1,\dots,a_N : \\ \sum_{j=1}^N a_j=1 ~ \text{and} \\ a_j > 0 ~\forall~ j = 1, \dots, n}}
\sum_{i=1}^d \sum_{j=1}^N \frac{\gamma_{ij}^2(\theta)}{a_j}
=
\argmin_{\substack{a_1,\dots,a_N : \\ \sum_{j=1}^N a_j=1 ~ \text{and} \\ a_j > 0 ~\forall~ j = 1, \dots, n}}
\sum_{j=1}^N \frac{\sum_{i=1}^d \gamma_{ij}^2(\theta)}{a_j}.
\end{equation*}
The method of Lagrange multipliers gives that
\begin{align*}
\ahat_j 
\propto
\left(\sum_{i=1}^d \gamma_{ij}^2(\theta)\right)^{1/2}
& =
\left(\sum_{i=1}^d  \left [ \sum_{t \in C_j} \nabla_i \left \{ \log \, p_\theta(Y_t \mid x_t) + \log \, p_\theta (x_t \mid x_{t-1}) \right \} \right ]^2\right)^{1/2};
\end{align*}
\cref{app.lagrange} contains a proof of this claim.
This completes the proof of the proposition.
\end{proof}
The weights determined by \cref{eq.optim.weight.known.latent} sum over all components of the parameter $\theta$. In the presence of a rare state, which we can assume to be state one without loss of generality, the weights \eqref{eq.optim.weight.known.latent} will be efficient in inferring parameters not associated with state one; this includes emission parameters for states other than state one, $\{\phi_k\}_{k=2}^K$, and transition parameters for states not involving state one, $\{A_{kk'}\}_{k,k'=2}^K$. However, they will not be efficient in inferring parameters associated with state one, as the presence of state one in the $n$-th subsequence is not going to significantly increase the weight $\ahat_n$. This suggests that using a single importance sub-sampling strategy for all components of $\theta$ is not efficient; instead we consider component-specific importance weights. By minimizing the variance of each component of the stochastic gradient estimate with component-specific weights, we attain a smaller sum of variances than a single importance weighting scheme, and thus expect better performance of the stochastic gradient MCMC scheme. In \cref{sec.grad.est}, we will revisit this issue and demonstrate that using the component-specific weighting scheme can achieve a lower root-mean-square error (RMSE) when estimating gradients.

Let $\{\ahat_n^{(i)}\}_{n=1}^N$ denote component-specific importance weights for $i=1,\dots,d$, where $a_n^{(i)} > 0$ and $\sum_{n=1}^N a_n^{(i)} = 1$. Then $\hatnabla_i \ell(\theta) = \{a_J^{(i)}\}^{-1} \sum_{t \in C_J} \nabla_i \{ \log \, p_\theta(Y_t \mid x_t) + \log \, p_\theta (x_t \mid x_{t-1}) \}$ for $J \sim \text{Multinomial}\{N, (a_1^{(i)},\dots,a_N^{(i)})\}$ is an unbiased estimator of $\hatnabla_i \ell(\theta)$. An unbiased estimator of the complete gradient $\nabla_\theta \ell(\theta)$ is thus given by $(\hatnabla_1 \ell(\theta) \cdots \hatnabla_{d} \ell(\theta) )^\top$. 
The following corollary to \cref{prop.imp.known.latent} minimizes $\EE [\{\hatnabla_i \ell(\theta) - \nabla_i\ell(\theta)\}^2]$ for $i=1,\dots,d$.

\begin{corollary}[Component-specific optimal importance weights for 
known latent states]
\label{cor.imp.known.latent.target}
$\EE [\{\hatnabla_i \ell(\theta) - \nabla_i\ell(\theta)\}^2]$, $i = 1, \dots, d$, is minimized for the targeted weights
\begin{equation}
\ahat_{n}^{(i)}
\propto
\left|\sum_{t \in C_n} \nabla_i \left \{ \log \, p_\theta(Y_t \mid x_t) + \log \, p_\theta(x_t\mid x_{t-1})\right \} \right|, ~~ n=1,\dots,N. 
\label{eq.optim.wts.targeted}
\end{equation}
\end{corollary}

\begin{proof}
The proof of \cref{cor.imp.known.latent.target} is very similar to the proof of \cref{prop.imp.known.latent}, with the only difference being that we omit the sum over the components of $\theta$.
\end{proof}

We thus propose sampling $d$ separate subsequences $\mb^{(i)} = \{J^{(i)},\dots,J^{(i)}_M\}$ for $i=1,\dots,d$, respectively, which are used to estimate individual components of the gradient. 
This results in improved estimates of the rare latent state parameters since subsequences without relevant information are omitted with high probability.

As an illustrative example, consider the case of $K$ one-dimensional Gaussian emission distributions with unknown means and variances. We focus on the emission parameters for the moment. Let $\phi_k = (\mu_k,\sigma^2_k)$ for $k=1,\dots,K$, where $\mu_k$ and $\sigma_k^2$ denote the mean and variance of the $k$th Gaussian emission distribution. In this setting,
\begin{align*}
\sum_{t \in C_n} \nabla_{\mu_k} \log \, p_\theta(Y_t \mid x_t) & = 
\sum_{t \in C_n}\left \{ \frac{ Y_t - \mu_k}{\sigma_k^2} \I(x_t = k) \right \}
=
\frac{c_{n,k}}{\sigma_k^2} (\ybar_{n,k} - \mu_k),
\\ 
\sum_{t \in C_n} \nabla_{\sigma_k^2} \log \, p_\theta(Y_t \mid x_t) 
& =
\sum_{t \in C_n}\left [ \left \{ -\frac{1}{2\sigma_k^2} + \frac{(Y_t - \mu_k)^2}{2\sigma_k^4}\right \} \I(x_t = k) \right ] 
=
\frac{c_{n,k}}{2\sigma_k^4}\bigg ( S_{n,k}^2-\sigma_k^2 \bigg ) ,
\end{align*}
where $\I(\cdot)$ denotes the indicator function, $c_{n,k} = \sum_{t\in C_n}\I(x_t = k)$ is the number of times latent state $k$ occurs in subsequence $C_n$, and $\nabla_{\mu_k}$ and $\nabla_{\sigma_k^2}$ denote gradients with respect to $\mu_k$ and $\sigma_k^2$, respectively.
Additionally, $\ybar_{n,k}= c_{n,k}^{-1} \sum_{t\in C_n} Y_t\I(x_t=k)$ is the sample mean of observations corresponding to latent state $k$ in subsequence $C_n$, and $S_{n,k}^2 = c_{n,k}^{-1}\sum_{t\in C_n} (Y_t-\mu_k)^2\I(x_t = k)$ is the corresponding sample variance assuming the true mean is $\mu_k$. For completeness, we define $\ybar_{n,k} = S^2_{n,k} = 0$ if $c_{n,k}=0$. 
We then obtain weights $\ahat_n^{(\mu_k)} \propto c_{n,k}|\ybar_{n,k}-\mu_k|$ and $\ahat_n^{(\sigma_k^2)} \propto c_{n,k} 
(\sigma_k^2 + S^2_{n,k})^{1/2}$ for the emission parameters, where we have elected to use the parameters themselves in the superscripts rather than specifying the location in $\theta$ for clarity.

Unfortunately, the optimal weights \eqref{eq.optim.wts.targeted} depend on the current parameter value $\theta$ and knowledge of the latent state. These issues make \cref{eq.optim.wts.targeted} difficult to use in practice as it would need to be recalculated at every MCMC iteration thereby increasing computational demands. Furthermore, any theoretical guarantees based on these results would be impossible to verify.  Instead, we will focus on leveraging \cref{eq.optim.wts.targeted} to design a targeted sub-sampling strategy. In \cref{sec.grad.est}, we will provide numerical evidence showing the improved performance using resulting weighting scheme. 

When $T$ is large, we assume the posterior concentrates around the maximum a posteriori estimate of $\theta$, hereby denoted by $\theta_\MAP$. This is a common assumption made while designing efficient posterior sampling schemes for independent observations \citep{baker2019control,bierkens2019zig}. While this ignores the multi-modality of the posterior arising due to label switching in the latent states in the case of hidden Markov models, this is inconsequential for predictive inference. 
We thus fix $\theta$ in \cref{eq.optim.wts.targeted} at $\theta_\MAP$, thereby removing the need to recalculate weights at every iteration. This leads to weights
\begin{equation}
\ahat_{n}^{(i)} 
\approx 
\atilde_n^{(i)} 
\propto 
\left|\sum_{t \in C_n} \nabla_{\theta_i} \left \{ \log \, p_\theta(Y_t \mid x_t) + \log \, p_\theta(x_t\mid x_{t-1}) \right \} \right |_{\theta = \theta_\MAP}.
\label{eq.optim.wts.heuristic}
\end{equation}

To keep things simple, we evaluate the maximum a posteriori estimate $\theta_\MAP$ assuming improper priors $p_0(\theta) \equiv 1$. In the Gaussian example, this gives rise to the weighting schemes $\atilde_n^{(\mu_k)} \propto c_{n,k}|\ybar_{n,k}-\ybar_k|$ and \blue{$\atilde_n^{(\sigma_k^2)} \propto c_{n,k} |S_{n,k}^2 - S^2_{k}|$}, where $\ybar_k = c_k^{-1} \sum_{t=1}^T Y_t \I(x_t = k),$ $S^2_{n,k} = c_{n,k}^{-1}\sum_{t \in C_n} (Y_t - \ybar_k)^2\I(x_t = k)$, and $S^2_{k} = c_{k}^{-1}\sum_{t=1}^T (Y_t - \ybar_k)^2\I(x_t = k)$. Full details of this derivation can be found in Appendix \ref{app:gmm_weights}. Thus for the $k$th emission mean and variance, subsequences are given larger weights if they have a high number of state $k$ observations or if the observations associated with state $k$ differ greatly from the average $\ybar_k$ of all state $k$ observations in the time series. Weighting schemes for other emission distributions are available in \cref{table:weights}.

Finally, we choose weights for the transition matrix as $\atilde_n^{( A_{kk'})} \propto \xi_{n,k,k'}$, where 
$\xi_{n,k,k'}$ is the number of transitions from $k'$ to $k$ in subsequence $n$, that is, $\sum_{t \in C_n} \I(x_{t-1}=k', x_t=k)$. 
The weights are largest for those subsequences having the highest number of transitions from state $k'$ to state $k$.

In this paper, we do not infer the initial distribution $p(X_0)$. While the inference of $p(X_0)$ can be an important problem in the setting of small $T$, the scope of this paper is on the setting of large $T$ and thus the influence of $p(X_0)$ becomes negligible. Moreover, $p(X_0)$ will typically not come into play as we use a buffering of size $B$ when updating other parameters, so that we are only doing a forward and backward pass within a stepsize of $B$ for every mini-batch.

\begin{table}
\centering 
\renewcommand{\arraystretch}{1.5}
\scalebox{0.9}{
\begin{tabular}{|c || c | c |}
\hline
Emission family & $\atilde^{(\mu_k)}_n \propto$ & $\atilde^{(\sigma^2_k)}_n \propto $\\
\hline
Gaussian & $c_{n,k}|\ybar_{n,k}-\ybar_k|$ & \blue{$c_{n,k} |S_{n,k}^2 - S^2_{k}|$} \\
\hline
Student's-t & $\displaystyle \left|\sum_{t \in C_n} \frac{(\nu+1)(Y_t-\ybar_k)/s_k^2}{\nu+(Y_t-\ybar_k)^2/s_k^2} \I(x_t = k)\right| $ &
$\displaystyle 		
\left|\sum_{t\in C_n} \frac{\nu[(Y_t-\ybar_k)^2-s_k^2]}{\nu s_k^4 + s_k^2(Y_t-\ybar_k)^2} \I(x_t=k) \right| $\\
\hline
Poisson & $c_{n,k}|\ybar_{n,k}-\ybar_k|$ & NA\\
\hline
\end{tabular}	
}
\caption{Weight table for different emission distribution families. (a) Gaussian: $\N(x; \mu,\sigma^2)= (\sigma \sqrt{2 \pi})^{-1} \exp[-\{(x-\mu)/\sigma\}^{2}/2]$; (b) Student's-t: $t(x;\mu,\sigma^2,\nu) = \Gamma\{(\nu+1)/2\}/ \{ \sqrt{\nu \pi} \sigma \Gamma(\nu/2) \} [1 + \{(x-\mu)/\sigma\}^2/\nu]^{-(\nu+1)/2}$; (c) Poisson: $\mathrm{Poisson}(x;\mu) = \mu^x \exp(-\mu)/x!$, where $x\in \NN$.}
\label{table:weights}

\end{table}

Unfortunately, the latent states are unknown in practice, which means that we cannot calculate the weights directly. We propose an additional approximation for this in the next section.

\subsection{Clustering observations to approximate importance weights}
\label{sec.clust}

To address dependence of the importance weights on the unknown latent states $X_0,\dots,X_T$, a natural approach is to rely on a rough initial estimate of the states. There are relevant existing algorithms, including 
the segmental $k$-means algorithm \citep{Juang1990} and a modification of the expectation-maximization (EM) algorithm \citep{Volant2014hidden} to address the sequential dependence of hidden Markov models. Both of these approaches rely on the forward-backward algorithm to handle the Markovian structure of the latent states and maximize the objective function 
\begin{equation}
\log\, p(\bY \mid \theta, \bX) 
=
p(X_0) \sum_{t=1}^T \left \{\log \, p_\theta(Y_t \mid X_t) + \log\, p_\theta(X_t\mid X_{t-1}) \right \}
\label{eq.latent.likelihood}
\end{equation}
over $\bX$ and $\theta$. A disadvantage of these approaches is that a lot of computation is spent on an initial estimate of the latent states, which is only used in the importance weighting.

As a faster and simpler alternative, we instead use $k$-means clustering applied to the observations; this is fast and is additionally parallelizable in the case of extremely large $T$ 
\citep{kraj2008parakmeans,kerdprasop2010parallelization}. While this algorithm ignores dependence in the latent states, when the observations arising from different states are well-separated, the $\log \, p_\theta(Y_t\mid X_t)$ terms will dominate \cref{eq.latent.likelihood}, so that $k$-means clustering can be expected to give reasonable latent state estimates. In the context of Gaussian emissions, the intuition here is akin to the connection between $k$-means clustering and the EM algorithm applied to the fitting of Gaussian mixture models (GMMs).
When the clusters are well-separated, 
the clustering assignments obtained by $k$-means tend to coincide with those from fitting a GMM. 

Regardless of the algorithm used to obtain an initial estimate of the latent states for use in importance weighting, the results are subject to ambiguity in label assignments.
As such, the resulting weighting scheme will target one of multiple equivalent modes of the posterior and one should not expect the resulting SGLD scheme to visit the other equivalent modes efficiently. 
Rather, up to label ambiguity, our TASS
algorithm is expected to converge more rapidly than current SGLD algorithms for HMMs, and additionally produce more accurate posterior summaries for rare state parameters and predictions of future rare state occurrence.

Consider a clustering of the observations $Y_1, \dots, Y_T$ into $K$ clusters $\Z_1,\dots,\Z_K$, and let the cluster label for $Y_t$ be $z_t \in \{1,\dots,K\}$. 
In this case, we replace the latent states in \cref{eq.optim.wts.heuristic} (which we assumed were known a priori) with our clustering based labels $\{z_t\}_{t=1}^T$ to obtain weights $\wt{\alpha}_n^{(i)}$.
These weights are then used as inputs in \cref{alg.sg.mcmc.hmm} wherein lines (2) and (3) are iterated over each component of the gradient. \blue{A MATLAB implementation of this method may be found at github.com/young1062/TASS.}  We now turn to numerical applications which demonstrate the improved performance of TASS relative to uniform sub-sampling.

\section{Synthetic data experiments} \label{sec.numerics}


\subsection{Single rare latent state hidden Markov model}
\label{sec.single.rare}

We first consider the situation where we have a single rare latent state. We consider Gaussian emission distributions with means $\mu_1, \dots, \mu_K$ and variances $\sigma_1^2, \dots, \sigma_K^2$ for the $K$ latent states, respectively. 
We set the true transition matrix $A$ to be 
\begin{align*}
A 
& =
\begin{pmatrix}
0.990 & 0.005 & 0.495 \\
0.005 & 0.990 & 0.495 \\
0.005 & 0.005 & 0.010 \\
\end{pmatrix},
\end{align*}
and set the means and variances of the emission distributions to be $\mu=(-20,0,20)^\top$ and $\sigma^2=(1,1,1)^\top$, respectively. The means are well-separated relative to the standard deviations; this is similar to the rare latent state dynamics presented in the real data example in \cref{sec.solar.flare}.

We simulate $2 \times 10^6$ observations from the model. We use the first $T = 10^6$ observations as our training dataset, and the latter $T_ {\text{test}} = 10^6$ observations as our test dataset.
The stationary distribution of the considered latent Markov chain is $(0.4975, 0.4975, 0.0050)^\top$. States one and two are common states, while state three is rare. 
We set the stepsize $\epsilon = 10^{-6}$ for both samplers (uniform and TASS). We set the buffer size $B=5$, the half-width $L=2$ and the sub-sample size per iteration $S=10$.
We consider fairly diffuse priors for the $\mu_k$s and $\sigma_k^2$s: $\mu_k \sim \N(0,10^2)$ and $\sigma_k^2 \sim \IG(3,10)$ for $k=1,\dots,K$, where $\IG$ denotes an inverse-gamma distribution. We also consider a $\mathrm{Dirichlet}(1,1,1)$ prior for each column of $A$. We impose the identifiability constraint $\mu_1 < \mu_2 < \mu_3$ on the parameter space by post-processing the posterior samples in order to tackle the label switching issue.

\begin{figure}
\centering
\begin{subfigure}[b]{0.45\textwidth}
\centering
\includegraphics[width=\textwidth]{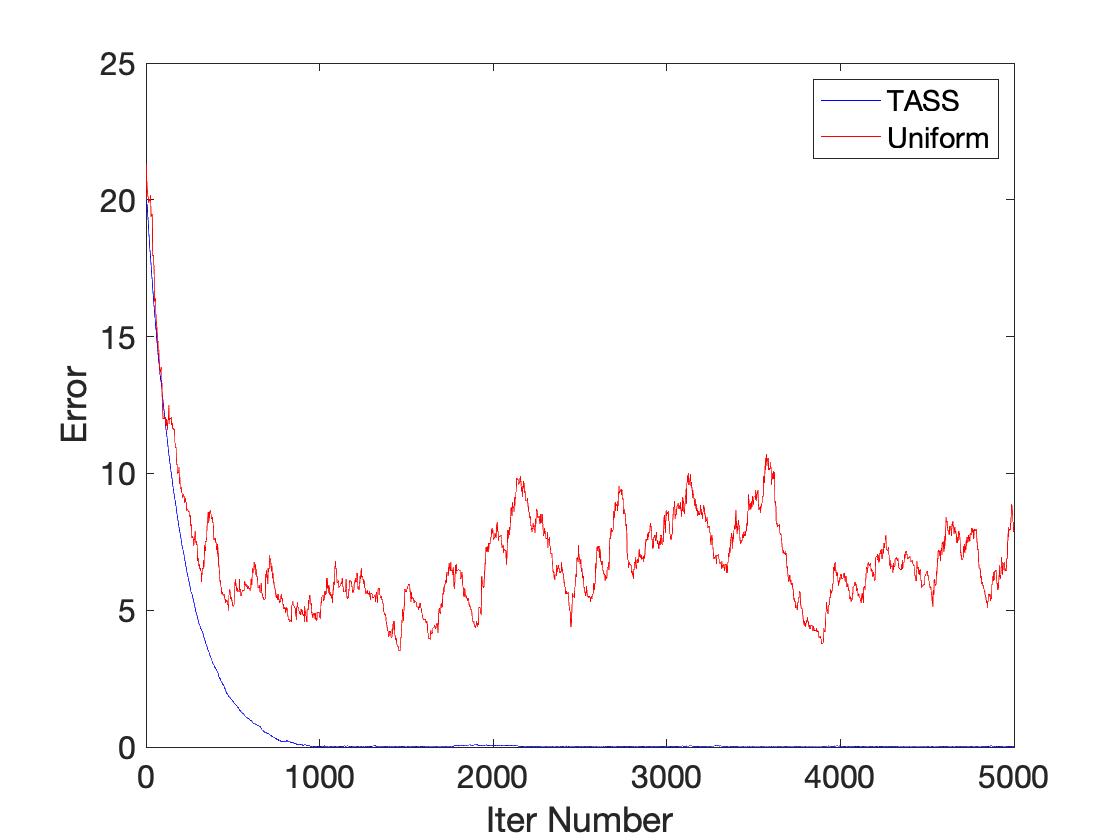}
\end{subfigure}
\hfill
\begin{subfigure}[b]{0.45\textwidth}
\centering
\includegraphics[width=\textwidth]{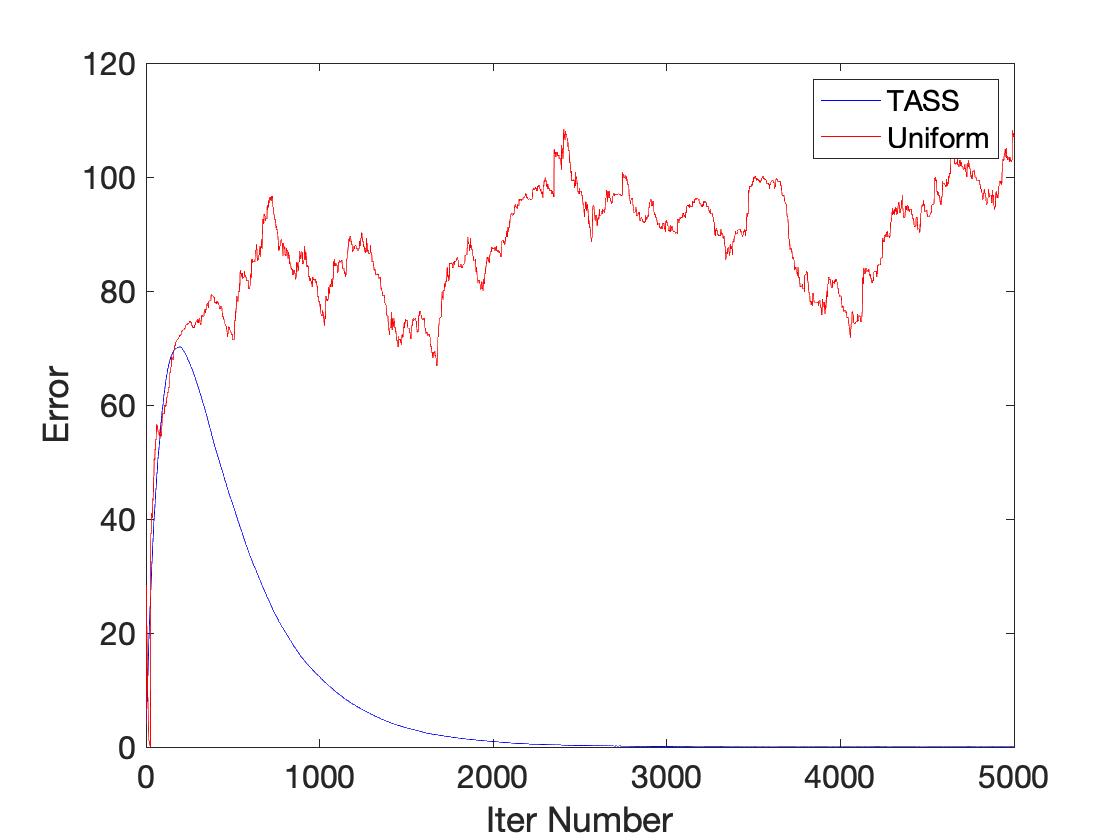}
\end{subfigure}
\caption{Distance between the true and sampled parameter values at different iterations. $|\sigma_3^2-\sigma^2_{3, \text{true}}|$ versus iteration number (left); $|\mu_3-\mu_{3, \text{true}}|$ versus iteration number (right).}
\label{fig:paramest}
\end{figure}

We first compare the performance of rare latent state parameter estimation. In this case, the rare latent state emission parameters are $(\sigma_3^2,\mu_3)$. We compare $|\sigma_3^2-\sigma^2_{3,\text{true}}|$ and $|\mu_3-\mu_{3,\text{true}}|$. The plots are shown in \cref{fig:paramest}. This indicates that TASS converges close to the true parameters in $2 \times 10^3$ iterations, while uniform sub-sampling appears to have substantial asymptotic (large number of MCMC samples) bias.

\begin{figure} 
\centering
\includegraphics[scale=0.25]{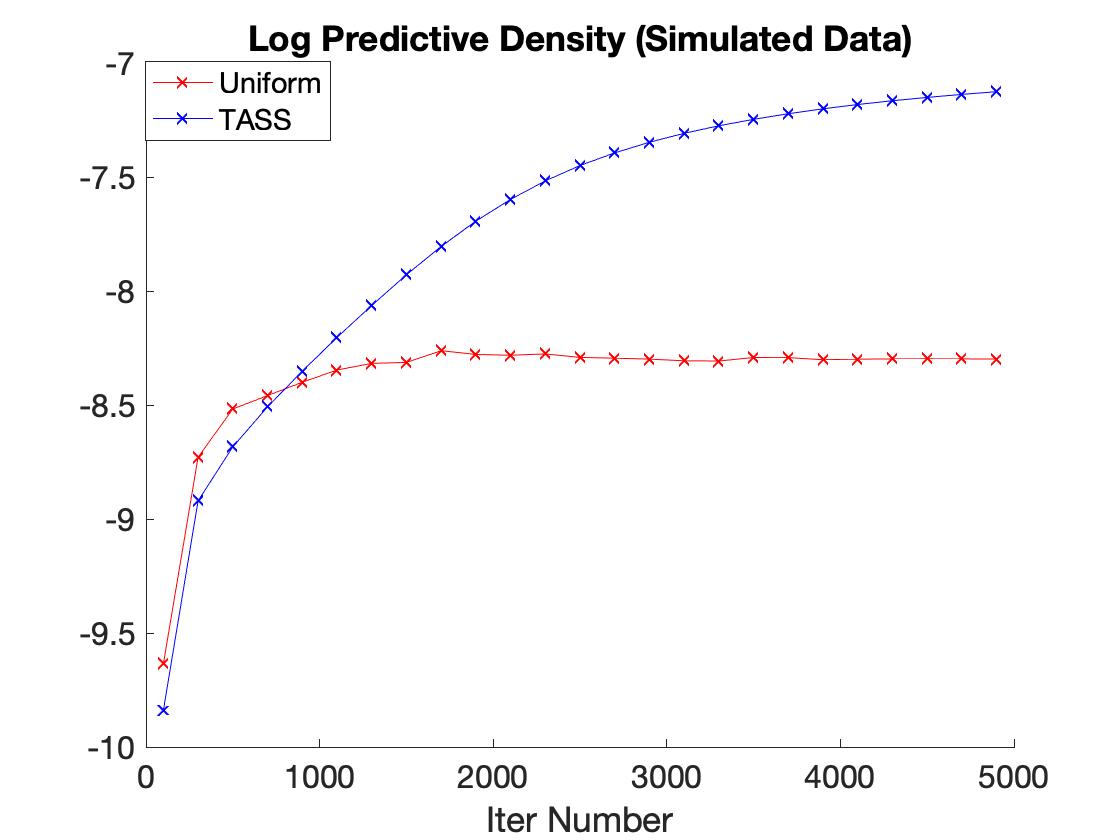}
\caption{Log predictive density of held-out data for one rare latent state example.}
\label{fig:simdatapred}
\end{figure}

We use predictive performance to compare different sub-sampling algorithms, relying on the expected log predictive density \citep{gelman2014understanding}. \blue{To focus on rare state prediction,} we hold out $n_{\text{hold}}=200$
uniformly sampled observations associated with \blue{ the rare state} without replacement from the test dataset, that is, $\{Y^{(1)}, \cdots, Y^{(n_{\text{hold}})}\} \sim \U(\{Y_t \mid X_t=3\})$. We then estimate the mean log predictive likelihood by
\begin{equation}\label{eq:heldout}
\frac{1}{n_{\text{hold}}}\sum_{r=1}^{n_{\text{hold}}} \log \left ( \frac{1}{Z} \sum_{z=1}^{Z} p ( Y^{(r)} \mid \theta_z, \blue{X^{(r)}=3} ) \right ).
\end{equation}
The log-predictive density \blue{of rare states} is plotted in \cref{fig:simdatapred}. We observe that the log predictive density of TASS is larger than that of the uniform sub-sampling strategy after $10^3$ iterations.
This indicates that TASS is able to predict rare data more accurately than SGLD with uniform sub-sampling.

\subsection{Multiple rare latent states} 
\label{sec.multiple.rare}

We consider multiple rare states in this section.
We set the transition matrix $A$ to be
\begin{equation*}
A = \begin{pmatrix}
0.9990 & 0.1000 & 0.1000 \\
0.0005 & 0.9000 & 0.0000 \\
0.0005 & 0.0000 & 0.9000 
\end{pmatrix}.
\end{equation*}
In this case, the stationary distribution of the latent states is $(0.990, 0.005, 0.005)^\top$ which indicates that both state two and three are rare latent states. We set $\mu=(0,-20,20)^\top$ and $\sigma^2 = (1,1,1)^\top$. We simulate $2 \times 10^6$ observations from the model. We use the first $T = 10^6$ observations as our training dataset, and the latter $T_ {\text{test}} = 10^6$ observations as our test dataset.

\begin{figure} 
\centering
\begin{subfigure}[b]{0.45\textwidth}
\centering
\includegraphics[width=\textwidth]{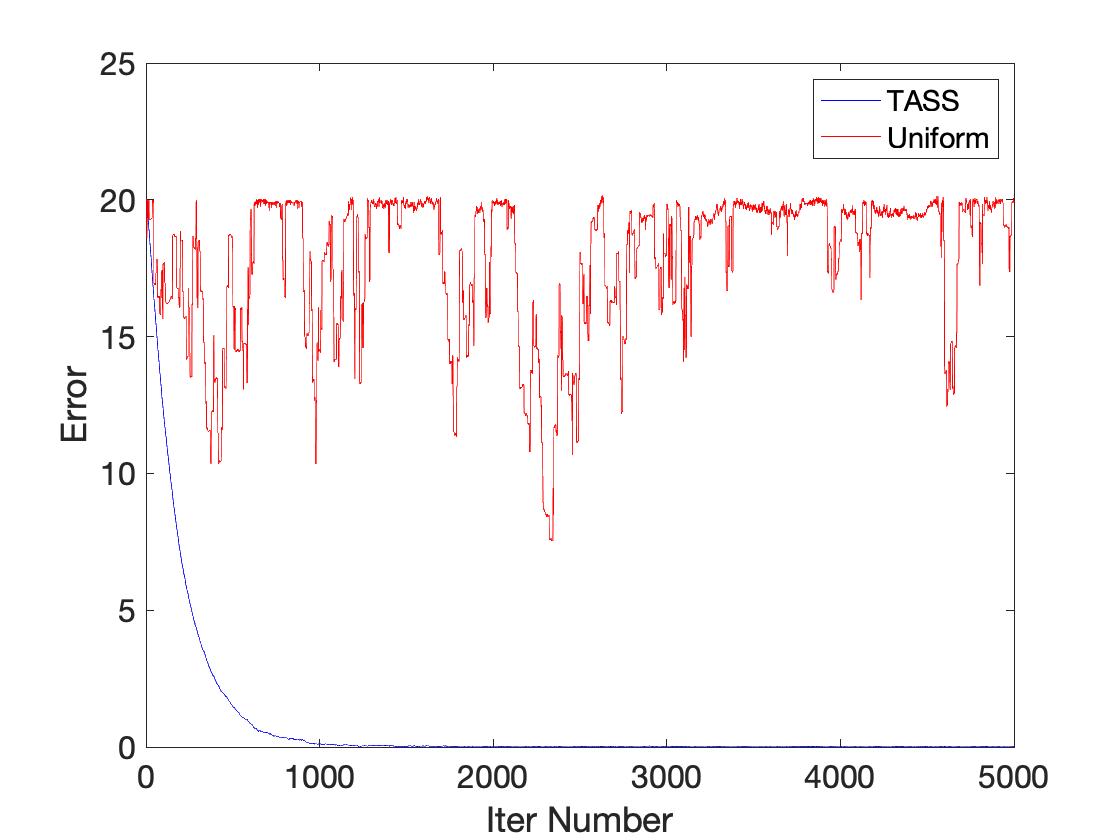}

\label{fig:mu2sim2rare}
\end{subfigure}
\hfill
\begin{subfigure}[b]{0.45\textwidth}
\centering
\includegraphics[width=\textwidth]{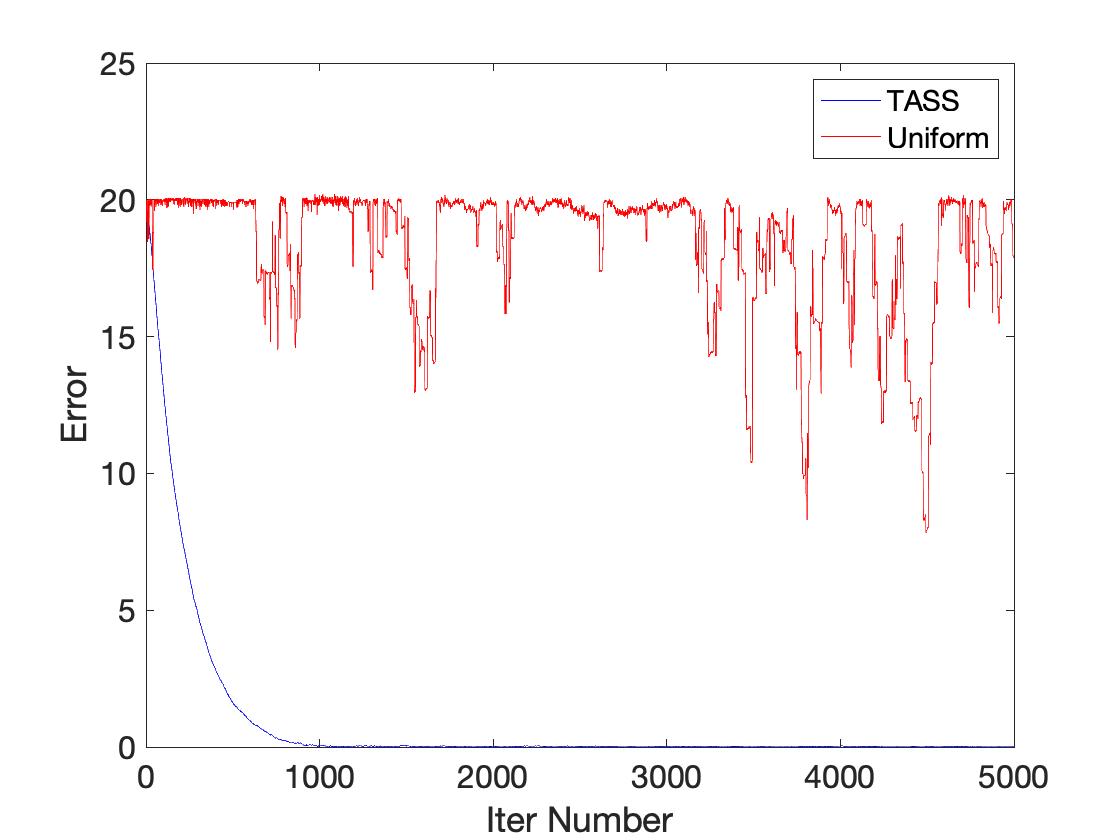}

\label{fig:mu3sim2rare}
\end{subfigure}
\caption{Distance between the true and sampled parameter values at different iterations when two rare latent states are present. $|\mu_2-\mu_{2, \text{true}}|$ versus iteration number (left); $|\mu_3-\mu_{3, \text{true}}|$ versus iteration number (right).}
\label{fig:paramest_2rare}
\end{figure}

Similar to \cref{sec.single.rare}, we compare the sub-sampling strategies by investigating their respective performance with respect to parameter estimation and predictive performance. To estimate the performance of parameter estimation, we compare $|\mu_2-\mu_{2,\text{true}}|$ and $|\mu_3 - \mu_{3,\text{true}}|$, that is, the distance between emission parameters associated with rare latent states at each iteration to its true value. The results are displayed in \cref{fig:paramest_2rare}, which indicates that TASS converges close to the truth in $ 10^3$ iterations, while uniform sub-sampling has substantial asymptotic bias.

\begin{figure} 
\centering
\begin{subfigure}[b]{0.45\textwidth}
\centering
\includegraphics[width=\textwidth]{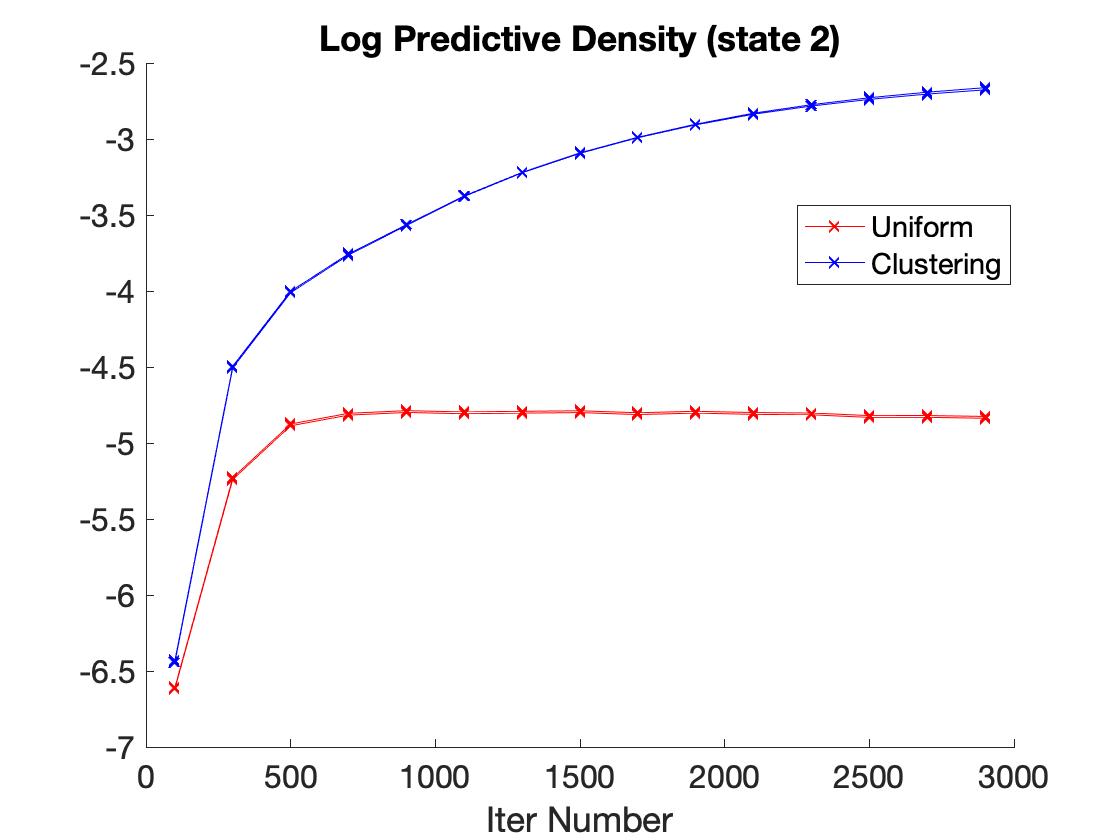}
\label{fig:sim2pred2}
\end{subfigure}
\hfill
\begin{subfigure}[b]{0.45\textwidth}
\centering
\includegraphics[width=\textwidth]{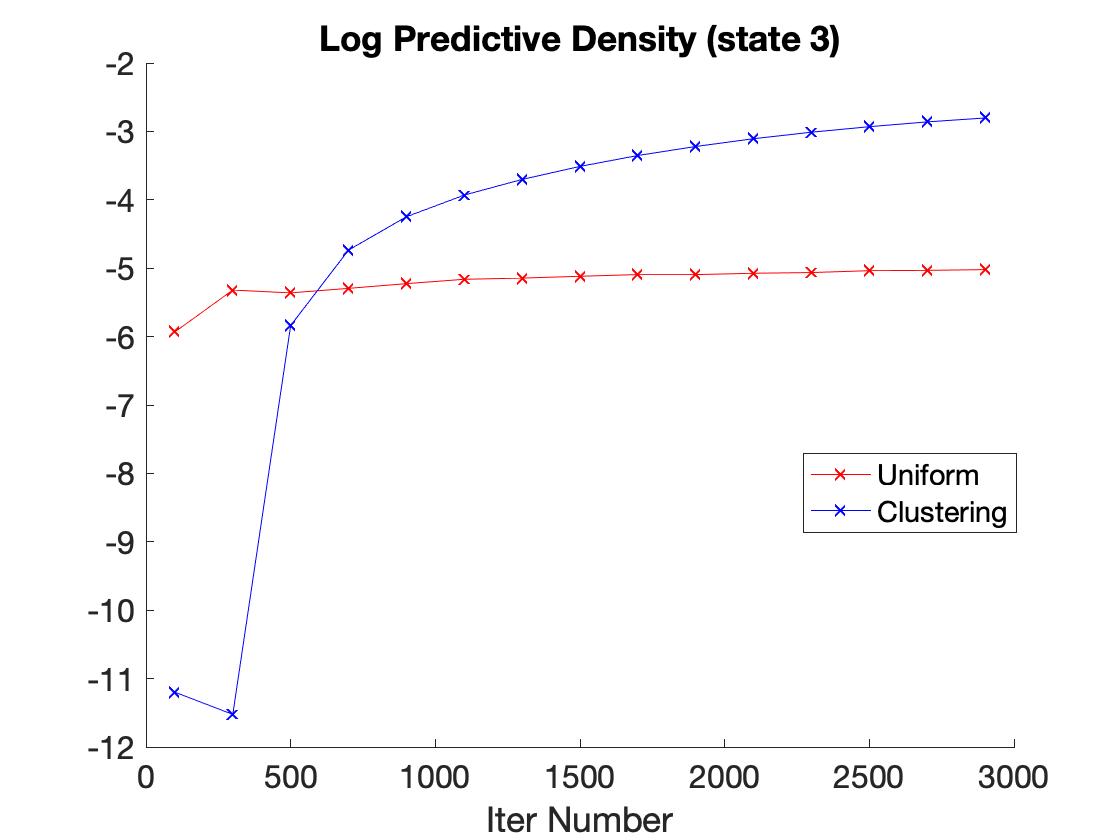}
\label{fig:sim2pred3}
\end{subfigure}
\caption{Log predictive density of held-out data for state two (left) and state three (right) when two rare latent states are present.}
\label{fig:sim2pred}
\end{figure}

In addition to estimation, we also examine the performance of predicting held-out rare spikes. We uniformly hold out $n_{\text{hold}}=200$ observations each from states two and three in a similar manner as described in the previous section. We then calculate the mean log predictive density as in \cref{eq:heldout}. The mean log predictive density versus iteration number is plotted in \cref{fig:sim2pred}, which again indicates that TASS achieves improved predictive performance relative to uniform sub-sampling at predicting rare states.

\subsection{No rare latent state} 
\label{sec.norare.state}
\begin{figure} 
\centering
\includegraphics[scale = 0.12]{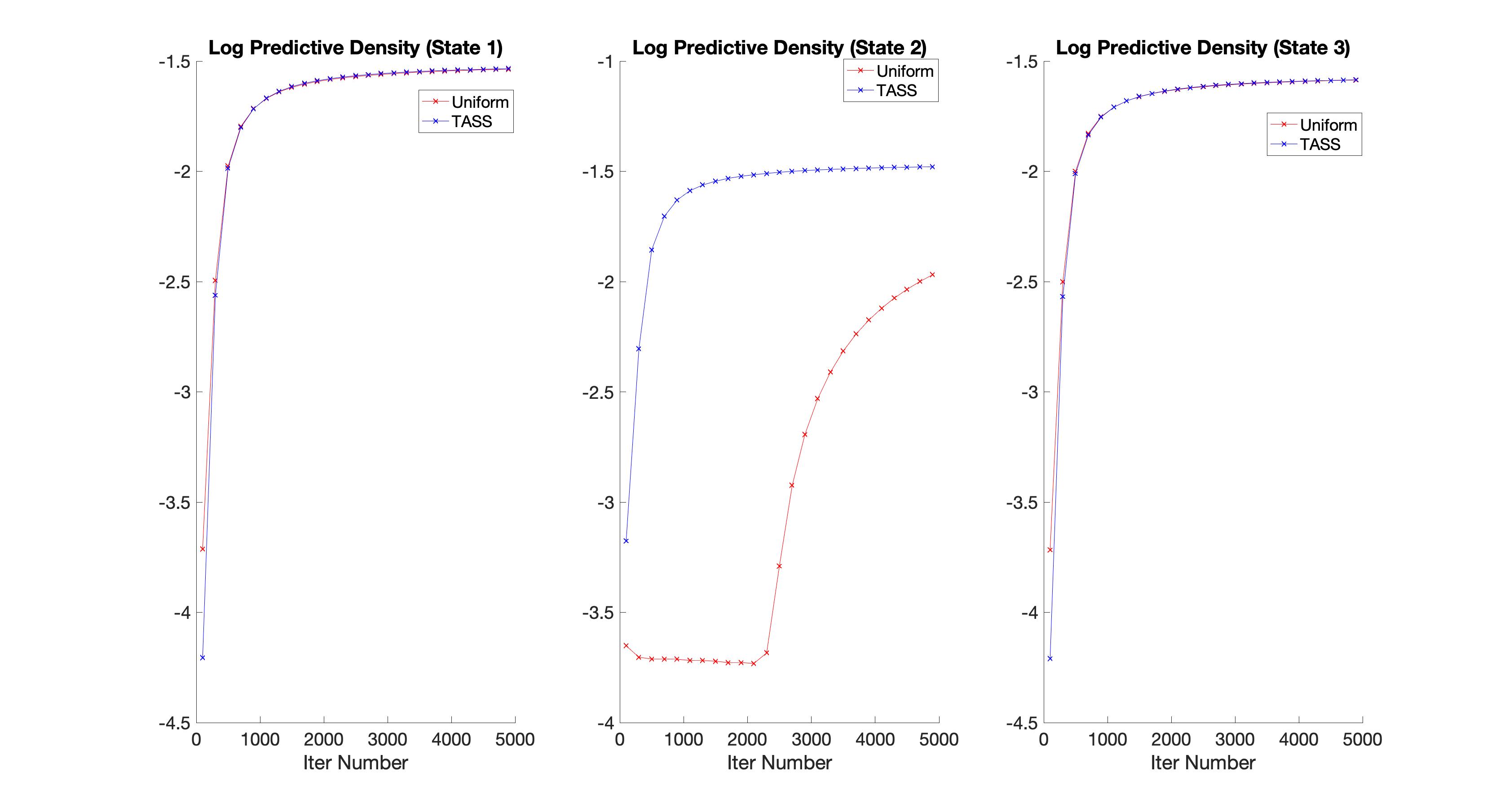}
\caption{Log predictive density of held-out data in the non-rare setting.}
\label{fig.simdata_pred_norare}
\end{figure}

The non-rare setting is not the main focus of our work. However, it is natural to wonder how TASS compares to uniform sub-sampling SGLD when all states are balanced. In this experiment, we set the true transition matrix $A$ to be 
\begin{align*}
A 
& =
\begin{pmatrix}
0.990 & 0.005 & 0.005 \\
0.005 & 0.990 & 0.005 \\
0.005 & 0.005 & 0.990 \\
\end{pmatrix},
\end{align*}
and set the means and variances of the emission distributions to be $\mu=(-20,0,20)^\top$ and $\sigma^2=(1,1,1)^\top$, respectively. The stationary distribution of the latent states is $\pi_0 = (1/3,1/3,1/3)^\top$. The prior distribution setting and the tuning parameters are the same as in \cref{sec.single.rare}. Similar to \cref{sec.single.rare}, we simulate $2 \times 10^6$ observations from the model. We use the first $T = 1 \times 10^6$ observations as our training dataset, and the latter $T_ {\text{test}} = 1 \times 10^6$ observations as our test dataset. We compare the log-likelihoods on held-out data in \cref{fig.simdata_pred_norare}. These two sub-samplers are close in predicting the held-out data from state $1$ and $3$, while we observe some gains for TASS in log-predictive density for state $2$, up to some variability in sampling and MCMC initialization.  We hypothesize that the computational gains are due to the nonuniform precomputed sampling weights for each mini-batch. Even without rare states, in updating the parameter $\theta_i$, the precomputation step of TASS will still assign higher weights for the mini-batches that contain the state corresponding to parameter $\theta_i$. Such non-uniformity makes TASS more targeted on the relevant mini-batches in subsampling, leads to gains in sampling efficiency, and hence gains in prediction. However, these improvements are larger 
in the presence of rare states.    

As we discussed in the previous paragraph, the sampling weights are not uniform across the entire time series in the non-rare setting.  However, compared to the rare state setting, the distribution of states is closer to uniform. In the presence of a single rare latent state (\cref{sec.single.rare}), the Kullback–Leibler divergence between the distribution of the weights for $\mu_3$ and the uniform distribution is $3.68$, whereas it drops to $1.08$ in the non-rare setting. 

\subsection{Run time analysis}
\label{sec.run.time.analysis}
We compare the run times of TASS and uniform sub-sampling based SGLD. 
As compared to uniform sub-sampling, TASS includes an additional clustering and weight calculation step.
This additional computational cost is independent of the number of MCMC iterations $n_\mcmc$. We highlight that the objective of this section is not to have a precise comparison of the performance of TASS versus that of uniform subsampling for a fixed computational cost as this will require both samplers to be written in an equivalently optimized form; the objective is rather to study the additional computational cost induced by TASS in exactly the same setup as the uniform sub-sampler. Hence we do not match the computational time of these two algorithms. We consider the same experimental setup as in \cref{sec.single.rare}
with increasing values of $T$ and display CPU run times in \cref{tab:runtime}, which shows the additional cost for TASS is small.

\begin{table}[ht]
\centering
\scalebox{0.9}{
\begin{tabular}{|c|cc|cc|cc|cc|}
\hline
& \multicolumn{2}{c|}{$n_{\mcmc} = 5 \times 10^3$} & \multicolumn{2}{c|}{$n_{\mcmc} = 6 \times 10^3$} & \multicolumn{2}{c|}{$n_{\mcmc} = 8 \times 10^3$} & \multicolumn{2}{c|}{$n_{\mcmc} = 10^4$}\\
\hline
& Uniform & TASS & Uniform & TASS & Uniform & TASS & Uniform & TASS \\
\hline
$T = 3\times 10^5$ & $190$ & $190$ & $238$ & $245$ & $283$ & $306$ & $353$ & $375$ \\
\hline
$T = 5\times 10^5$ & $249$ & $279$ & $284$ & $323$ & $378$ & $409$ & $492$ & $517$ \\
\hline
$T = 8\times 10^5$ & $330$ & $416$ & $390$ & $483$ & $525$ & $613$ & $638$ & $746$\\
\hline
$T = 10^6$ & $396$ & $536$ & $476$ & $648$ & $661$ & $845$ & $800$ & $986$ \\
\hline
\end{tabular}}
\caption{Run time comparison including k-means and the weight calculation step. All experiments were run on a 2.6 GHz CPU and all run times are CPU times measured in seconds.}
\label{tab:runtime}
\end{table}

\subsection{Comparison of posterior estimates with methods using full data}
\label{sec.comparison.full.mcmc}

\begin{figure}
\centering
\includegraphics[scale=0.2]{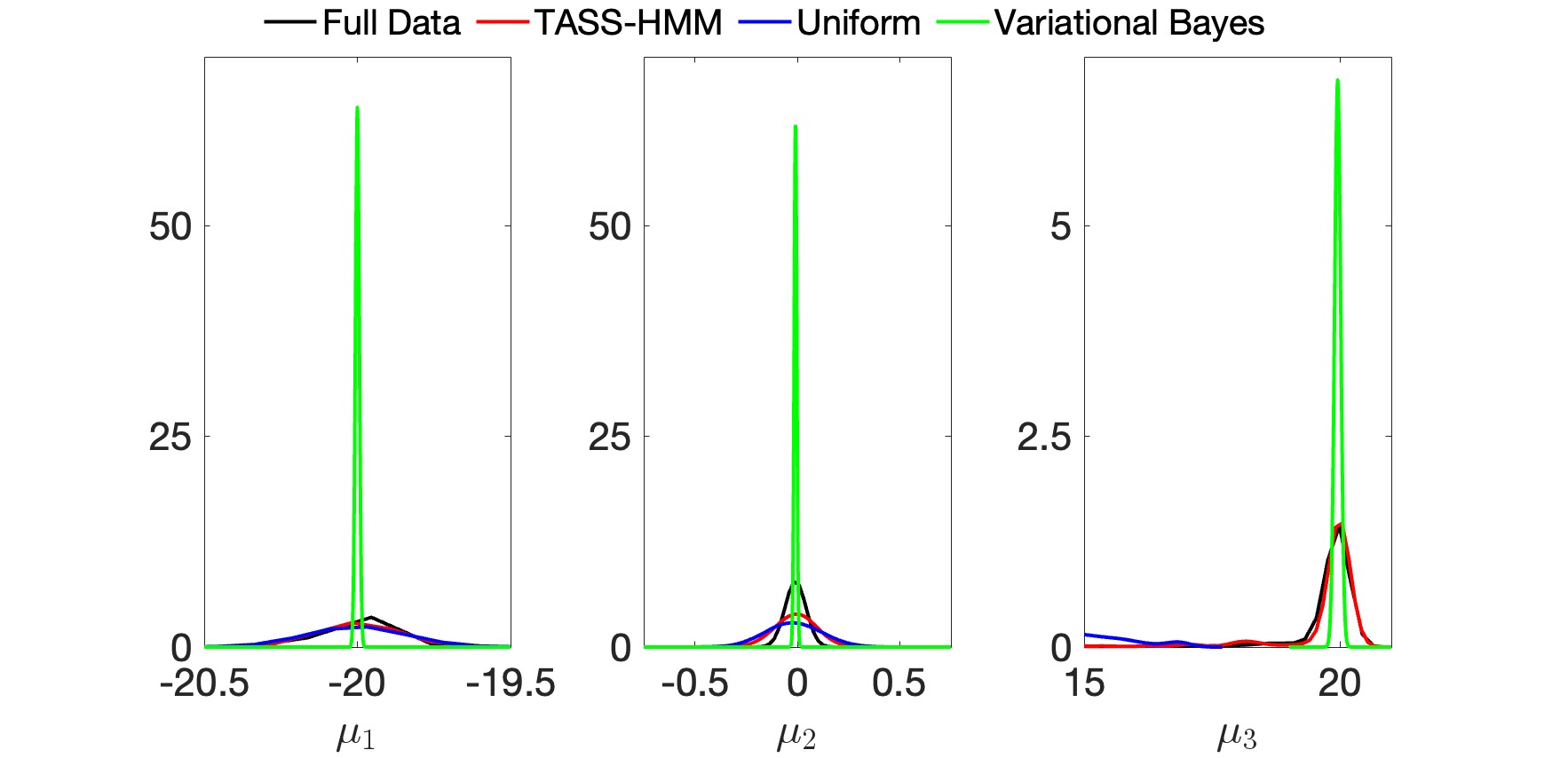}
\caption{Marginal posterior density estimates for the emission means generated by uniform sampling, TASS-HMM, full data Langevin dynamics, and full data Variational Bayes. 
For the rare state mean ($\mu_3$), uniform subsampling has a posterior density centered around 12, which we cut-off to more clearly show the performance of TASS-HMM compared to the methods using the full data.}
\label{fig:posterior_comparison}
\end{figure}

\blue{So far we have focused on predictive performance, but it is also important to consider uncertainty quantification. As such, we compare posterior estimates of TASS-HMM and uniform subsampling with alternative methods using the full data. We follow the experimental design in \cref{sec.single.rare} but reduce the sample size to $T = 10^5$ to make the implementation of MCMC without subsampling computationally feasible. 

In Figure \ref{fig:posterior_comparison}, we provide kernel density estimates of the marginal posterior distribution of the emission means using uniform sampling and TASS-HMM (25000 samples after 5000 burn-in). Additionally, we provide kernel density estimates based on samples generated using Langevin dynamics with the full data (5000 samples after 500 burn-in) with a step size of $10^{-5}$ along with optimal Gaussian densities obtained via Variational Bayes (VB). The VB densities were estimated using the BayesPy package \citep{bayespy}.   

For the common states ($\mu_1$ and $\mu_2$), all methods have posterior estimates centered around the true parameter values. However, VB consistently underestimates uncertainty relative to the full data MCMC in these settings, which is a known issue for VB. Both TASS and uniform subsampling match the full data uncertainty quite well for $\mu_1$ but slightly overestimate uncertainty for $\mu_2$, though TASS-HMM is incrementally better than uniform subsampling.  This discrepancy in posterior uncertainty for $\mu_2$ is an interesting question for future research.  

The more notable observations, which provides insight on the improved predictive performance of rare states in \cref{sec.single.rare}, are the posterior estimates for $\mu_3$.  As we observe in the rightmost column of Fig. \ref{fig:posterior_comparison}, uniform subsampling is extremely biased generating no samples near the true value of 20. We elected to cutoff the figure as shown to better compare the performance near the true value of 20.  Again, VB matches the mode well, but greatly underestimates uncertainty. Conversely, TASS-HMM matches the full data results exceptionally well, demonstrating its superior performance in the estimation of rare state parameters.}

\subsection{Gradient estimation analysis}
\label{sec.grad.est}

Finally, we compare the accuracy of the estimated gradient while using TASS to that obtained using a single weighting or uniform subsampling.
In particular, we compute the root-mean-square error (RMSE) of gradient estimates of the rate state mean under different sets of $(T, L)$ and different locations within the parameter space. We follow the experimental setup in \cref{sec.single.rare} except that we vary the choice of $T \in \{10^4, 5\times 10^4, 10^5\}$ and $L \in \{2, 12\}$. We fix $B = 5$ as we do not observe any significant difference when $B \geq 5$. Moreover, we let the rare state mean $\mu_3$ vary between $0$ to $3$ standard deviations from its truth $\mu_{3,\text{true}}$, while fixing other parameters on their respective true values. The relatively low value of $T$ permits us to evaluate the true gradient without subsampling. We first randomly draw $n_{\mathrm{rep}} = 10^3$ mini-batches with replacement for both TASS and the uniform subsampler. Then we compare their RMSE in \cref{tab.mse.grad}. 

The RMSE of the gradient estimates obtained from TASS is significantly lower than its counterpart obtained from uniform subsampling. For gradient estimates near the true value, TASS and a single weighting behave comparably. Unlike TASS, however, the performance of a single weighting degrades quickly as we move away from the true rare state value.  When $\delta_{\mu,3} = |\mu_3- \mu_{3,\text{true}}| = 3\sigma$, that is, the current parameter is far from the truth, the performance of TASS as measured by RMSE is a two-three fold improvement over the single weighting scheme.  

\begin{table}[ht]
\centering
\begin{tabular}{|c|cccccc|}
\hline
& \multicolumn{6}{c}{$L=2$} \vline \\
\hline
&  \multicolumn{3}{c|}{$T=10^4$} &  \multicolumn{3}{c}{$T = 10^5$} \vline\\
\hline
&  TASS & Single & \multicolumn{1}{c|}{Uniform} &  TASS & Single &  Uniform \\
\hline
$\delta_{\mu,3} = 0$ &  $4.9  \times 10^1$ & $5.9\times 10^1$ & \multicolumn{1}{c|}{$1.7  \times 10^3$} & $4.8\times 10^2$ & $6.7 \times 10^2$  & $1.7  \times 10^3$
\\
$\delta_{\mu,3} = \sigma$ & $4.9  \times 10^1$ & $8.1\times 10^1$ & \multicolumn{1}{c|}{$1.7  \times 10^3$} & $4.8 \times 10^2$ & $9.1 \times 10^2$ & $1.7  \times 10^3$
\\
$\delta_{\mu,3} = 2\sigma$ & $4.9  \times 10^1$ & $1.2\times 10^2$ & \multicolumn{1}{c|}{$1.7  \times 10^3$}  & $4.8 \times 10^2$ & $1.3 \times 10^3$ &  $1.7  \times 10^3$
\\
$\delta_{\mu,3} = 3\sigma$ & $4.9  \times 10^1$ & $1.6\times 10^2$ & \multicolumn{1}{c|}{$1.7  \times 10^3$} & $4.8 \times 10^2$ & $1.9 \times 10^3$  & $1.7  \times 10^3$
\\ 
\hline
\end{tabular}
\begin{tabular}{|c|cccccc|}
\hline
& \multicolumn{6}{c}{$L=12$} \vline \\
\hline
&  \multicolumn{3}{c|}{$T=10^4$} &  \multicolumn{3}{c}{$T = 10^5$} \vline\\
\hline
&  TASS & Single & \multicolumn{1}{c|}{Uniform} &  TASS & Single &  Uniform \\
\hline
$\delta_{\mu,3} = 0$ &  $4.9  \times 10^1$ &$5.4  \times 10^1$ & \multicolumn{1}{c|}{$1.7  \times 10^3$} & $4.7\times 10^2$ & $5.9  \times 10^2$ & $1.7  \times 10^3$
\\
$\delta_{\mu,3} = \sigma$ & $4.9  \times 10^1$ & $6.4  \times 10^1$ &\multicolumn{1}{c|}{$1.7  \times 10^3$}  & $4.7 \times 10^2$ & $7.5  \times 10^2$ & $1.7  \times 10^3$
\\
$\delta_{\mu,3} = 2\sigma$ & $4.9  \times 10^1$ & $8.6  \times 10^1$ & \multicolumn{1}{c|}{$1.7  \times 10^3$} & $4.7 \times 10^2$ & $1.1  \times 10^3$ &  $1.7  \times 10^3$
\\
$\delta_{\mu,3} = 3\sigma$ & $4.9  \times 10^1$ &$1.1  \times 10^2$  & \multicolumn{1}{c|}{$1.7  \times 10^3$} & $4.7 \times 10^2$ & $1.4  \times 10^3$ & $1.7  \times 10^3$
\\ 
\hline
\end{tabular}
\caption{RMSE of the gradient estimates of the rate state mean under TASS, single, and uniform weighted subsampler under different half-width $L$, sample size $T$, and distance from the truth $\delta_{\mu,3} = |\mu_3- \mu_{3,\text{true}}|$ (for simulated data).}
\label{tab.mse.grad}
\end{table}

\section{Real data analysis} \label{sec.real.data}

\subsection{Solar flare data}
\label{sec.solar.flare}

Solar flares are sudden outbursts of energy from a small area of the sun's surface. Extreme solar flares can pose a great threat to satellites and thus it is crucial to predict such extreme flares. Solar flares exhibit a normal behavior most of the time with intermittent moments with occasional dangerous spikes. The increasing prevalence of sensor technology has made it possible to attain minute-to-minute data over many days resulting in massively long time series. In analyzing such massive time series, uniform sub-sampling schemes are likely to miss the rare dynamics, thereby hindering prediction and characterization of flare phenomena.

In this section, we consider solar flare X-ray data obtained from the GOES-13 and GOES-15 satellites from September 2017 to April 2018\footnote{The data are available at \url{https://www.ngdc.noaa.gov/stp/satellite/goes/index.html}.}. The dataset consists of $T_ {\text{total}} = 347875$ time points. Observations are taken from two different satellites, and we average these two different observations prior to our HMM analyses. When one of the satellites is missing an observation at a time point, we simply rely on the other satellites data at that time.
Previous work suggests that some hidden switching regimes exist behind the observable X-ray flux \citep{stanislavsky2019solar,stanislavsky2020prediction}.

\begin{figure}[ht]
\centering
\begin{subfigure}[b]{0.495\textwidth}
\centering
\includegraphics[width=\textwidth]{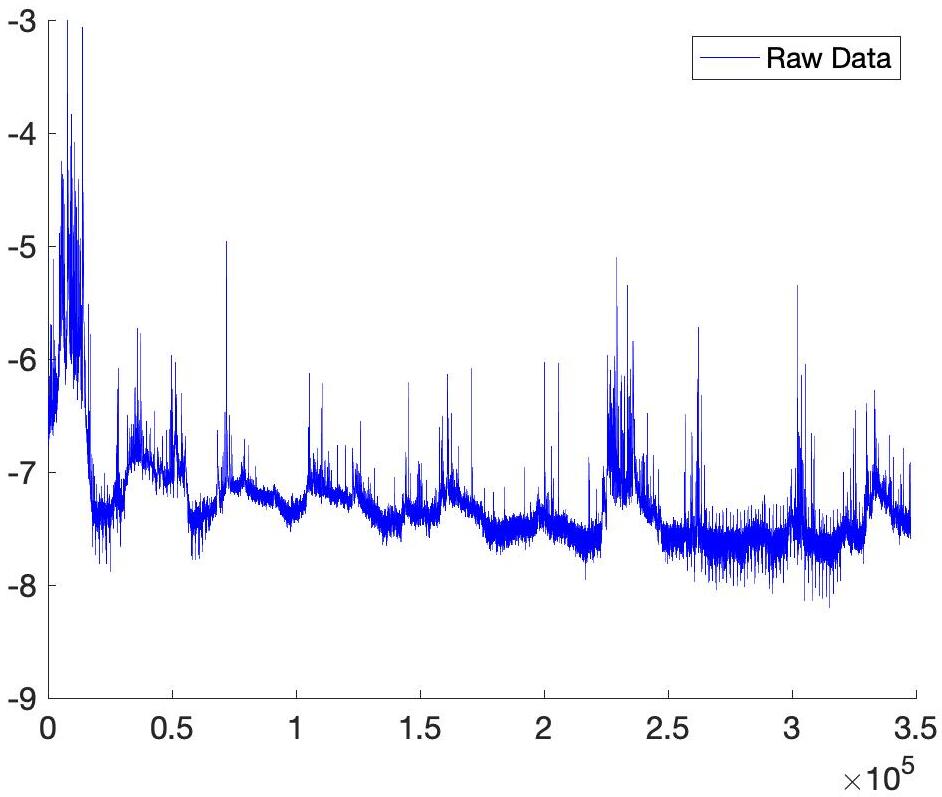}
\caption{Raw data.}
\end{subfigure}
\hfill
\begin{subfigure}[b]{0.495\textwidth}
\centering
\includegraphics[width=\textwidth]{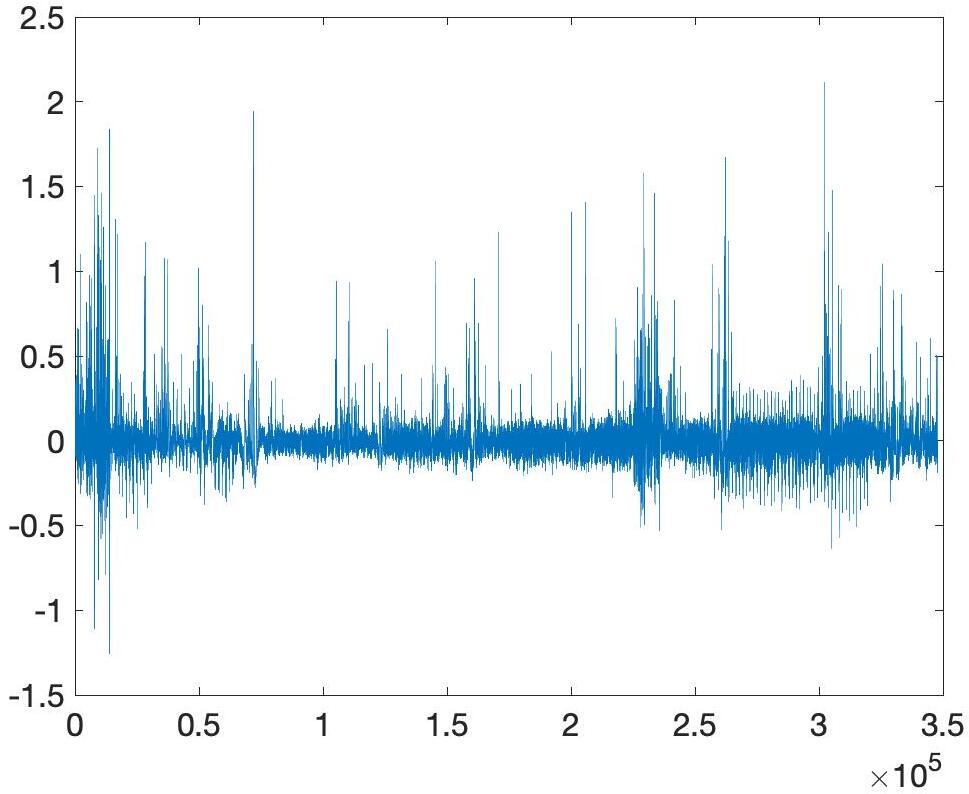}
\caption{Preprocessed data.}
\end{subfigure}
\caption{The raw and pre-processed solar flare data with the latter obtained after subtracting a smooth baseline.}
\label{fig:data}
\end{figure}

We plot the observed series on the log-scale in \cref{fig:data}(a), which clearly suggests that this series is non-stationary. We use a piece-wise quadratic baseline to fit the time trend, and then focus our Bayesian modeling on the residual process after subtracting the baseline. 
We choose the baseline knots adaptively to adjust for differing volatility at different times. In particular, we place fifty knots evenly in $[1, 1.86\times10^4]$, fifty knots in $[1.9\times 10^4, 2.2\times 10^5]$, fifty knots in $[2.2\times 10^5, 2.5\times 10^5]$, and fifty knots in $[2.5\times 10^5, T_{\text{total}}]$; the knots are evenly-spaced within each interval. The residual series after baseline subtraction is plotted in \cref{fig:data}(b), where we observe occasional rare spikes. Our main objective is to infer the parameters characterizing the rare dynamics and predict the rare spikes. We set $T=3\times10^5$ and use the first $T$ observations as our training dataset and the remaining observations as our test dataset.

\begin{figure}[ht]
\centerline{\includegraphics[width=1.25\textwidth]{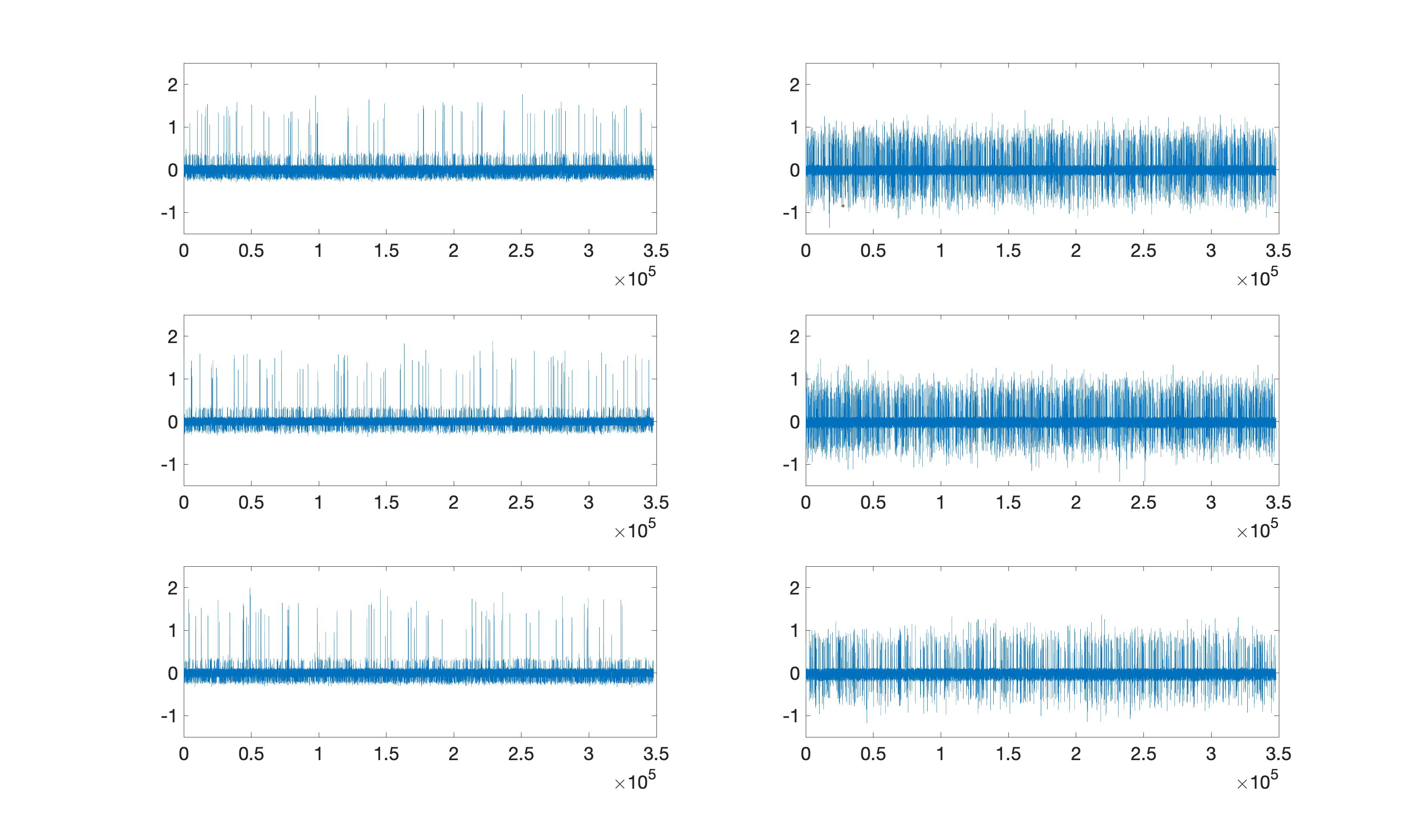}}
\caption{Predicted series plots for solar flare data. (\textit{top left}), (\textit{middle left}) and (\textit{bottom left}) are predicted series based on the $4000$th, $4500$th, and $5000$th sample of SGLD using TASS, respectively; (\textit{top right}), (\textit{middle right}) and (\textit{bottom right}) are predicted series based on $4000$th, $4500$th, and $5000$th sample of SGLD with uniform sub-sampling, respectively. 
}
\label{fig:predseries}
\end{figure}

We consider a model with $K=4$ latent states because solar flares can be classified into four categories (Class B, C, M, X) according to their strength as pointed out in \cite{stanislavsky2020prediction}. We assume Gaussian emission distributions and consider the same priors as in \cref{sec.single.rare}. We sub-sample the data by setting half-width $L=2$, sub-sample size $S=10$, and buffer size $B=5$. We run SGLD with $5 \times 10^3$ iterations for TASS and uniform sub-sampling. The tuning of $L$, $S$, and $B$ involves trade off between the accuracy of gradient estimates and computational cost per iteration, with details presented in \cref{sec.parameter.tuning}.

To qualitatively compare the two sub-sampling strategies, we compare the predicted series generated from both samplers, and favor the sampler that renders a predicted series similar to the original data. 
We plot the predicted series from both samplers based on the $(4000,4500,5000)$th posterior sample, respectively, in \cref{fig:predseries}, and compare them to the original data. We observe that uniform sub-sampling fails to capture the spike behavior, while the predicted series using TASS is able to mimic the rare spikes of the original data. Moreover, the transition rates between the rare latent state and the common states can also be mimicked using TASS.

\begin{figure}[ht]
\centering
\includegraphics[scale=0.2]{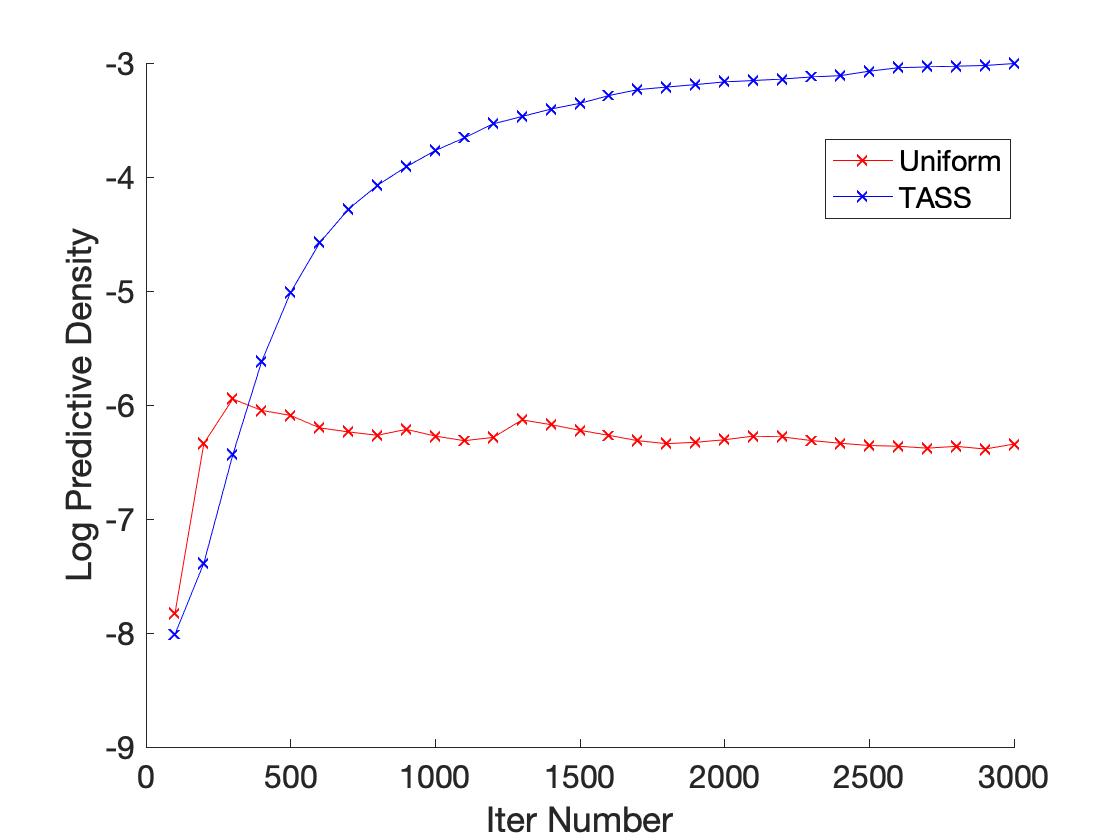}
\caption{Log predictive likelihood of the held-out rare spikes of the solar flare data.}
\label{fig.lppd.coverage}
\end{figure}

To quantitatively compare the performance of samplers, we set one as the threshold of rare spikes. We identify points that start exceeding this threshold in the test dataset and hold out ten subsequent observations for every point as the held-out dataset. We calculate the predictive likelihood for every held-out subsequence and plot it in \cref{fig.lppd.coverage}. We observe that TASS out-performs uniform sub-sampling after approximately $5 \times 10^2$ MCMC iterations with regard to the predictive performance of the rare spike.

\subsection{Sleep cycle data} \label{sec.sleep.cycle}

Sleep cycles between five distinct stages: wake, N1, N2, N3, and REM \citep{carskadon2005normal}. Due to these distinct latent states, HMMs have been widely used in modeling of sleep dynamics. It has become common to monitor sleep with high frequency sampling technology, leading to routine collection of massive datasets. The size of such datasets prevents a substantial barrier to Bayesian analyses. 

\begin{figure} 
\centering
\includegraphics[scale=0.12]{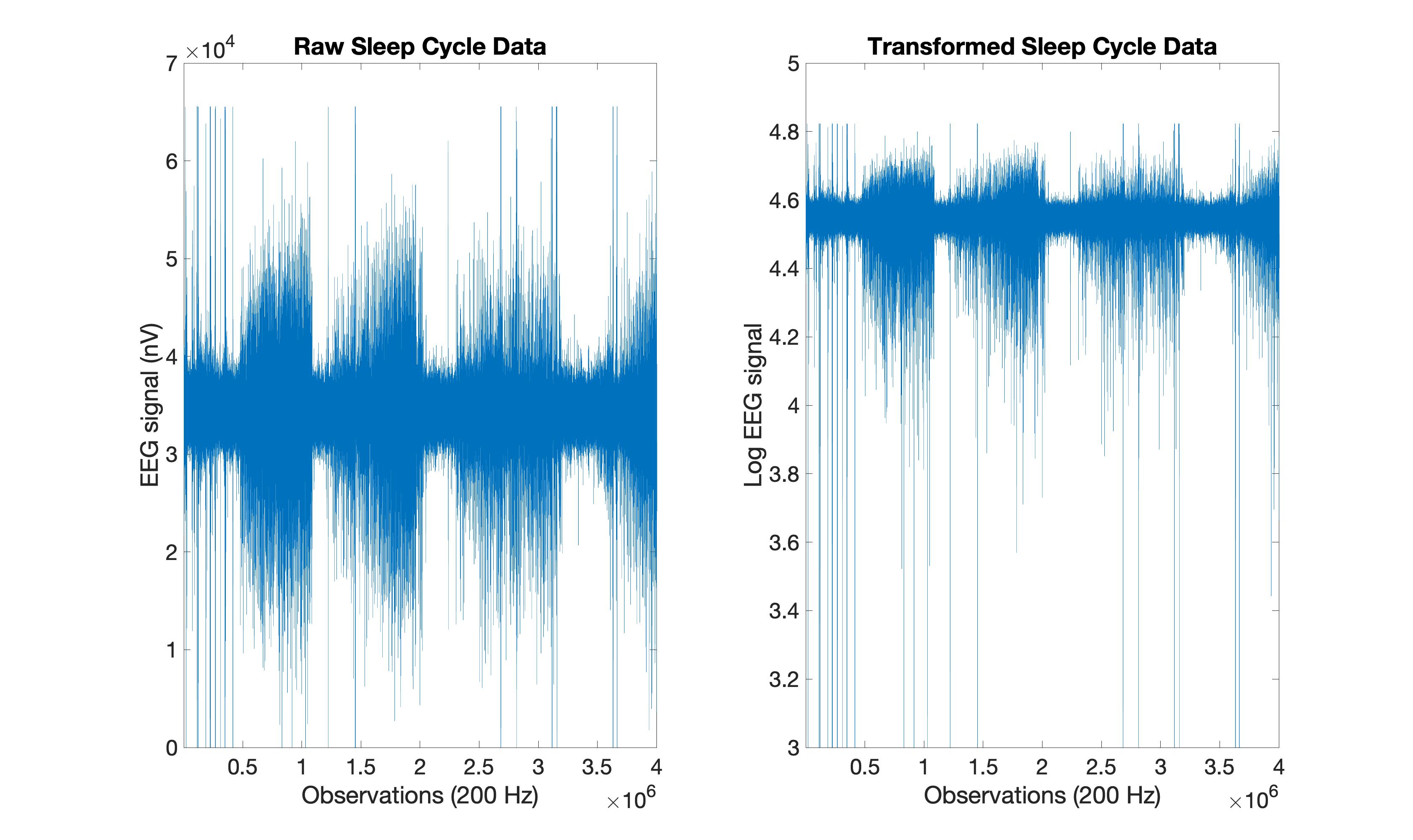}
\caption{The raw sleep cycle data (left) and the log-transformed sleep cycle data (right).}
\label{fig.data.sleep}
\end{figure}

We consider a Bayesian HMM for the ICASSP sleep dataset, which records electroencephalogram (EEG) data on brain activity. Previous work modeling single-channel sleep EEG data using HMMs include \cite{doroshenkov2007classification, ghimatgar2019automatic}. We use the C3A2 channel EEG recording from a single patient, measured in nanovolts and sampled at 200 Hz. We plot the raw data and the log-transformed data in \cref{fig.data.sleep}, where we see some rare spikes in the data. We fit an HMM on the log-transformed data. These spikes may correspond to an interesting brain state not captured by the typical four sleep state dynamics, or alternatively may represent measurement errors due to head motion. In either case, the rare spike state is an interesting target of further study. While further analysis of the mechanism of these rare spikes is a potentially interesting area of study, answering this question is beyond the scope of this paper and we only seek to identify the rare dynamics in the dataset.

\begin{figure} 
\centering
\includegraphics[scale=0.22]{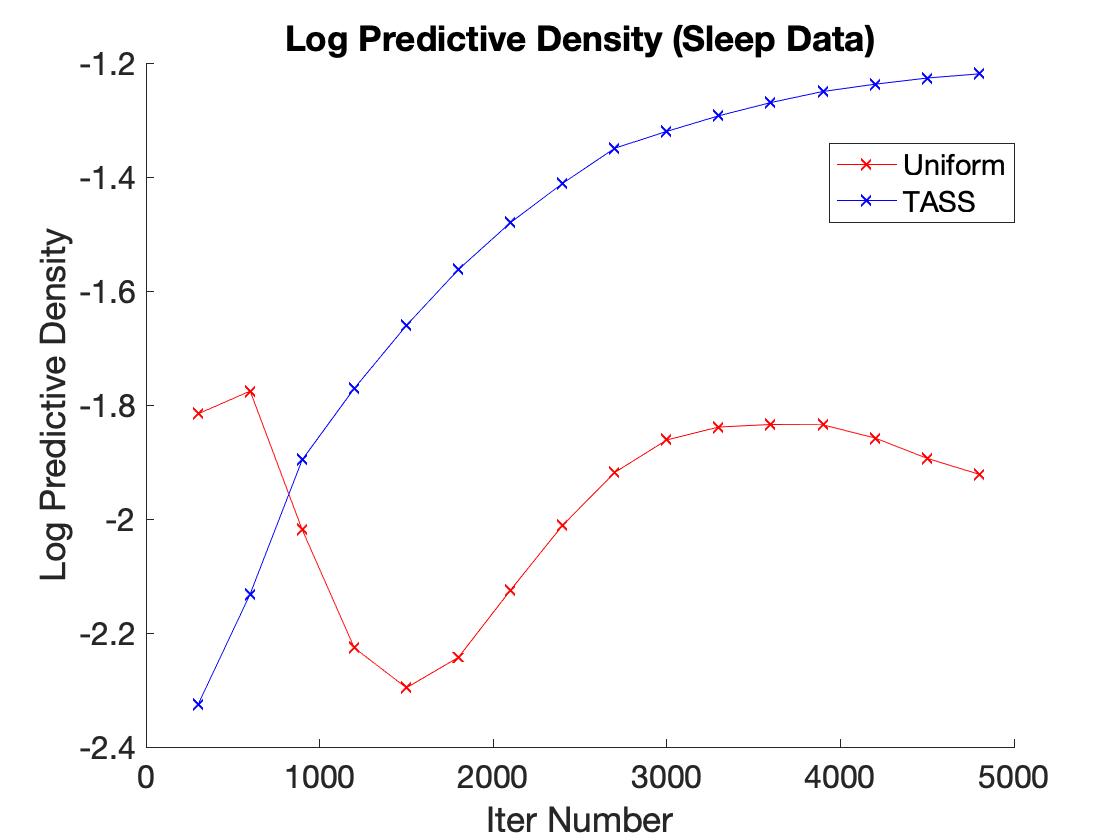}
\caption{Log predictive likelihood of the held-out rare spikes of sleep cycle data.}
\label{fig.lppd.sleep}
\end{figure}

We set $T_{\mathrm{train}} = 10^6$ and use the first $T_{\mathrm{train}}$ observations as the training dataset. We use observations $T_{\mathrm{train}}, \dots, 2 T_{\mathrm{train}}$ as our test dataset. We implement SGLD with $5 \times 10^3$ iterations for TASS and uniform sub-sampling on the training dataset. We set the number of states $K=4$. We also use grid-search to determine our hyperparameters half-width $L$, buffer size $B$ and sub-sample size $S$. The details of hyperparameter tuning are available in \cref{sec.parameter.tuning}. 
We use $L=2$, $S=5$, and $B=10$ as we do not observe sensitivity to these hyperparameters. 
After we obtain posterior samples, we hold out the rare spikes whose value is less than or equal to $3.8$ in the test dataset. We compare the log predictive performance on such held-out observations in \cref{fig.lppd.sleep}, which indicates that TASS dominates uniform sub-sampling regarding the predictive likelihood after $1000$ iterations.

\section{Discussion} \label{sec.discussion}

We have proposed a targeted sub-sampling (TASS) approach for stochastic gradient MCMC for discrete state-space hidden Markov models with a particular focus on improving rare latent state inference, which in turn leads to improved predictive performance. We have mainly focused on the case in which the states have well-separated emission distributions. The poorly-separated case presents substantial inferential difficulties for any posterior computation algorithm, and TASS is no exception. When components are substantially overlapping, the initial clustering algorithm that TASS leverages on may be inaccurate; indeed, theoretical guarantees on performance of clustering algorithms typically requires minimum separation conditions \citep{vempala2004spectral, yan2017convergence,zhao2020statistical}. Using the dynamics of the process can be helpful in better identifying the clusters. One possibility is to devote more computing resources to the initial clustering estimates; for example, by applying a preliminary non-Bayesian HMM algorithm that lacks uncertainty quantification.

We have elected to focus on MCMC algorithms based on overdamped Langevin dynamics to simplify the development of the material; however, TASS can be used within stochastic gradient Hamiltonian Monte Carlo \citep{chen2014stochastic} or stochastic gradient underdampled Langevin dynamics as well. These approaches would be better suited to handle the local modes inherent to HMM models. Additional techniques to speed up convergence such as reparameterization and preconditioning can also be combined straightforwardly with TASS \blue{and may provide better uncertainty quantification when there are large difference in posterior uncertainty}. While we did not discuss these explicitly, the former was used in estimating the emission parameters using the expanded mean reparameterization of \cite{patterson2013stochastic}.

Moreover, moving beyond Monte Carlo algorithms, the improved stochastic gradient estimates can be used within any inference scheme for hidden Markov models, including both Bayesian and frequentist approaches. In particular, it can be used within stochastic gradient descent and stochastic variational inference \citep{foti2014stochastic}. 
Furthermore, our coordinate-specific weighting scheme also contributes to the growing body of literature on the advantages of non-uniform data sub-sampling.

\clearpage
\bibliographystyle{ba}
\bibliography{references}

\section*{Acknowledgement}
DS and DD acknowledge support from National Science Foundation grant 1546130. DS acknowledges support from grant DMS-1638521 from SAMSI.

\appendix

\section{Additional calculations}

\subsection{Derivation of Weights for Gaussian Emissions}
\label{app:gmm_weights}
Assuming the latent states, $\{x_t\}_{t=1}^T$, were known, we introduced the following notation in \cref{sec.method}:
\begin{align*}
c_{n,k} 
& =
\sum_{t\in C_n} \I(x_t = k), 
\\
\ybar_{n,k} 
& =
\frac{1}{c_{n,k}}\sum_{t\in C_n} Y_t \I(x_t = k),
\\
s_{n,j}^2 
& =
\frac{1}{c_{n,k}}\sum_{t\in C_n} (Y_t - \mu_k)^2 \I(x_t = k),
\end{align*}
which represent the number, sample mean, and sample variance of state $k$ observations in the $n$th subsequence, respectively. Note that in the sample variance calculation we are using a known mean $\mu_k$.  We adopted the convention $\ybar_{n,k} = s_{n,k}^2 = 0$ when $c_{n,k}=0.$ These quantities can be related to the corresponding full data statistics on state $k$ observations through the following relationships.
\begin{align*}
c_k 
& = 
\sum_{t=1}^T \I(x_t = k) = \sum_{n=1}^N c_{n,k}, 
\\
\ybar_k 
& =
\frac{1}{c_k}\sum_{t=1}^T Y_t \I(x_t=k)= \frac{1}{c_k}\sum_{n=1}^N c_{n,k}\ybar_{n,k},
\\
S_k^2 &= \frac{1}{c_k}\sum_{t=1}^T (Y_t-\mu_k)^2 \I(x_t=k)=  \frac{1}{c_k}\sum_{n=1}^N c_{n,k}s_{n,k}^2.
\end{align*}
Assuming Gaussian emissions with unknown means $\{\mu_{j}\}_{j=1}^K$ and variances $\{\sigma_j^2\}_{j=1}^K$, the log-likelihood of the model parameters is
\begin{align*}
\ell(\theta) 
& = 
\sum_{t=1}^T\sum_{j=1}^K\bigg[-\frac{(Y_t - \mu_j)^2}{2\sigma_j^2} - \frac{1}{2}\log(\sigma_j^2)+ \sum_{l=1}^K\log(A_{jl})\I(x_{t-1} = l)\bigg] \I(x_t=j)  \\
& =
\sum_{n=1}^N \sum_{t \in C_n} \sum_{j=1}^K\bigg[-\frac{(Y_t - \mu_j)^2}{2\sigma_j^2}- \frac{1}{2}\log(\sigma_j^2) + \sum_{l=1}^K\log(A_{jl})\I(x_{t-1} = l)\bigg] \I(x_t=j) \\
& = 
-\sum_{n=1}^N\sum_{j=1}^K \frac{c_{n,k}}{2}\bigg\{\frac{s_{n,k}^2}{\sigma_j^2} + \log(\sigma_j^2)\bigg\} + \sum_{n=1}^N\sum_{t\in C_n}\sum_{j=1}^k\sum_{l=1}^K \log(A_{jl})\I(x_t = j)\I(x_{t-1}= l).
\end{align*}
From the previous identities, we arrive at the (sum of) standard Gaussian likelihood(s) for each emission mean and variance
\begin{align*}
\ell(\theta)
& = 
f(A,\bX)-\sum_{j=1}^K \frac{c_j}{2}\bigg\{\frac{S_j^2}{\sigma_j^2} + \log(\sigma_j^2)\bigg\}.
\end{align*}
Here we have compressed the transition parameters into the function 
\begin{align*}
f(A,{\bf X})=\sum_{n=1}^N\sum_{t\in C_n}\sum_{j=1}^k\sum_{l=1}^K \log(A_{jl})\I(x_t = j)\I(x_{t-1}= l).
\end{align*}
Thus, 
\begin{align*}
\nabla_{\mu_j} \ell(\theta) 
& =
\frac{1}{\sigma_j^2}\sum_{t=1}^T (Y_t - \mu_j)\I(x_t = k) = \frac{c_j}{\sigma_j^2}(\ybar_j-\mu_j) = \sum_{n=1}^N \frac{c_{n,j}}{\sigma_j^2}(\ybar_{n,j}-\mu_j),
\\
\nabla_{\sigma_j^2} \ell(\theta) &=\frac{c_j}{2}\bigg( \frac{S_j^2}{\sigma_j^4} - \frac{1}{\sigma_j^2}\bigg) = \frac{1}{2}\sum_{n=1}^N \bigg(\frac{c_{n,j}S_{n,j}^2}{\sigma_j^4} - \frac{c_{n,j}}{\sigma_j^2}\bigg).
\end{align*}
If we select subsequence $C_n$ with probability $\alpha_n$, then the weighted gradient $$\widehat{\nabla}_{\mu_j} \ell(\theta) =  \frac{1}{\alpha_n}\frac{c_{n,j}}{\sigma_j^2}(\ybar_{n,j}-\mu_j)$$
is an unbiased estimator of $\nabla_{\mu_j} \ell(\theta).$ Observe that 
\begin{align*}
E[\widehat{\nabla}_{\mu_j} \ell(\theta)]=\sum_{n=1}^N \alpha_n \bigg[\frac{1}{\alpha_n}\frac{c_{n,j}}{\sigma_j^2}(\ybar_{n,j}-\mu_j)\bigg] 
& =
\nabla_{\mu_j}\ell(\theta).   
\end{align*}
A similar calculation shows that the weighted stochastic gradient estimate
\begin{align*}
\widehat{\nabla}_{\sigma_j^2} 
& =
\frac{1}{2\alpha_n} \bigg(\frac{c_{n,j}S_{n,j}^2}{\sigma_j^4} - \frac{c_{n,j}}{\sigma_j^2}\bigg)
\end{align*}
is an unbiased estimator of $\nabla_{\sigma_j^2} \ell(\theta).$
We now seek separate probability weights $\alpha_i^{(\mu_j)}$, $j=1,\dots,n$ and 
$\alpha_i^{(\sigma_j^2)}$, $j=1,\dots,n$ which minimize the respective variances of each estimator.  We begin with the weights for $\mu_j$ suppressing the superscript on $\alpha_i$.  Note the variance of $\widehat{\nabla}_{\mu_j} \ell(\theta)$ is 
\begin{align*}
\var(\widehat{\nabla}_{\mu_j} \ell(\theta)) 
& =
\sum_{n=1}^N \alpha_n \bigg \{ \frac{1}{\alpha_n}\frac{c_{n,j}}{\sigma_j^2}(\ybar_{n,j}-\mu_j) - \frac{c_j}{\sigma_j^2}(\ybar_j-\mu_j)\bigg \}^2 
\\
& =
\sum_{n=1}^N  \bigg \{ \frac{c_{n,j}^2}{\alpha_n\sigma_j^2}(\ybar_{n,j}-\mu_j)^2 - 2\frac{c_{n,j}c_j}{\sigma_j^4}(\ybar_{n,j}-\mu_j)(\ybar_j - \mu_j) + \frac{\alpha_n c_j^2}{\sigma_j^4}(\ybar_j - \mu_j)\bigg \},
\end{align*}
so that 
\begin{align*}
\frac{\partial \var(\widehat{\nabla}_{\mu_j}\ell(\theta))}{\partial \alpha_m} 
& =
\frac{1}{\sigma_j^4}\bigg\{c_j^2(\ybar_j-\mu_j)^2-\frac{c_{m,j}^2}{\alpha_m^2}(\ybar_{m,j}-\mu_j)^2\bigg\},
\\
\frac{\partial^2 \var(\widehat{\nabla}_{\mu_j}\ell(\theta))}{\partial \alpha_m\partial \alpha_n} &= \frac{2c_{m,j}^2}{\alpha_m^2}(\ybar_{m,j}-\mu_j)^2 \delta_{mn}.
\end{align*}
It then follows that weights 
$$\alpha_m^{(\mu_j)} \propto \sqrt{c_{m,j}^2(\ybar_{m,j}-\mu_j)^2} = c_{m,j}|\ybar_{m,j}-\mu_j|$$
minimize the variance of the estimator.
Turning to $\widehat{\nabla}_{\sigma_j^2} \ell(\theta)$, we have
\begin{align*}
\var(\widehat{\nabla}_{\sigma_j^2} \ell(\theta)) 
& =
\sum_{n=1}^N \alpha_n\bigg( \frac{1}{2\alpha_n}\bigg[\frac{c_{n,j}S_{n,j}^2}{\sigma_j^4}-\frac{c_{n,j}}{\sigma_j^2}\bigg] - \frac{1}{2}\bigg[\frac{c_jS_j^2}{\sigma_j^4}- \frac{c_j}{\sigma_j^2}\bigg]\bigg)^2 \\
&=\frac{1}{4}\sum_{n=1}^N \bigg(\frac{1}{\alpha_n}\bigg[\frac{c_{n,j}S_{n,j}^2}{\sigma_j^4}-\frac{c_{n,j}}{\sigma_j^2}\bigg]^2 + \alpha_n\bigg[\frac{c_jS_j^2}{\sigma_j^4}- \frac{c_j}{\sigma_j^2}\bigg]^2 \\
&\hspace{3cm}- 2 \bigg[\frac{c_{n,j}S_{n,j}^2}{\sigma_j^4}-\frac{c_{n,j}}{\sigma_j^2}\bigg]\bigg[\frac{c_jS_j^2}{\sigma_j^4}- \frac{c_j}{\sigma_j^2}\bigg]\bigg),
\end{align*}
so that 
\begin{align*}
\frac{\partial \var(\widehat{\nabla}_{\mu_j}\ell(\theta))}{\partial \alpha_m} 
& =
\frac{1}{4}\bigg(\bigg[\frac{c_jS_j^2}{\sigma_j^4}- \frac{c_j}{\sigma_j^2}\bigg]^2 - \frac{1}{\alpha_n^2}\bigg[\frac{c_{m,j}S_{m,j}^2}{\sigma_j^4}-\frac{c_{m,j}}{\sigma_j^2}\bigg]^2\bigg)
\\
\frac{\partial^2 \var(\widehat{\nabla}_{\mu_j}\ell(\theta))}{\partial \alpha_m\alpha_n} 
& =
\frac{1}{2\alpha_n^3}\bigg[\frac{c_{m,j}S_{m,j}^2}{\sigma_j^4}-\frac{c_{m,j}}{\sigma_j^2}\bigg]^2\delta_{mn}.
\end{align*}
Thus, optimal weights for the estimator of $\widehat{\nabla}_{(\sigma_j^2}\ell(\theta)$ are
\begin{align*}
\alpha_n^{(\sigma_j^2)}\propto \sqrt{\bigg[\frac{c_{m,j}S_{m,j}^2}{\sigma_j^4}-\frac{c_{m,j}}{\sigma_j^2}\bigg]^2} 
& \propto
c_{m,j}|S_{m,j}^2 - \sigma_j^2|.
\end{align*}
The final step in the derivation of the weights is to replace $\mu_j$ and $\sigma_j$ with their \emph{maximum a posteriori} estimates. Assuming improper priors, this is is equivalent to using the MLE estimates $\ybar_j$ and $S_j^2$ for $\mu_j$ and $\sigma_j^2$ respectively.

\subsection{Proposition required in the proof of \texorpdfstring{\cref{prop.imp.known.latent}}{} } \label{app.lagrange}

\begin{proposition} 
The following is true:
\begin{align*}
\argmin_{\substack{a_1,\dots,a_N : \\ \sum_{j=1}^N a_j=1 ~ \text{and} \\ a_j > 0 ~\forall~ j = 1, \dots, n}} \sum_{i=1}^n \frac{x_i^2}{a_i} 
& = 
\left ( \frac{|x_1|}{\sum_{i=1}^n |x_i|}, \dots, \frac{|x_n|}{\sum_{i=1}^n |x_i|} \right ).
\end{align*}
\end{proposition}

\begin{proof}
The proof uses the method of Lagrange multipliers. The Lagrangian for this problem is given by:
\begin{align*}
L(a_1, \ldots, a_n, \lambda) = \sum_{i=1}^n \frac{x_i^2}{a_i} + \lambda \left(1 - \sum_{i=1}^n a_i \right),
\end{align*}
where $\lambda$ is the Lagrange multiplier. 
Setting the partial derivatives of $L$ with respect to each variable to zero gives:
\begin{align*}
-\frac{x_i^2}{a_i^2} + \lambda = 0 \quad \text{for } i = 1, \ldots, n \quad \text{and} 
\quad 
1 - \sum_{i=1}^n a_i = 0.
\end{align*}
The first set of equations gives $\lambda = x_i^2/a_i^2 \implies a_i = |x_i|/\lambda$. Substituting this into the second equation $\sum_{i=1}^n a_i = 1$ gives $\sum_{i=1}^n |x_i|/\lambda = 1 \implies \lambda = \sum_{i=1}^n |x_i|$. Thus $a_i = |x_i|/(\sum_{j=1}^n |x_j|)$.
\end{proof}

\section{Parameter tuning}
\label{sec.parameter.tuning}

The choice of buffer length $B$, the number of mini-batches used per iteration $S$, and the half-length of each mini-batch $L$ can affect the bias and variance of the gradient estimate per iteration, and hence the performance of the subsampling algorithm. As analyzed in \cite{aicher2019stochastic}, the bias of the buffered gradient estimator scales $\OO(\rho_{\theta}^B)$, where $0 < \rho_{\theta} < 1$ is determined by the spectral gap of the HMM. Regarding $S$ and $T$, the mean square error of the gradient estimator scales as $\OO\left( 1/SL \right )$. The computational cost per iteration is $\OO((L+B)S)$. Therefore, increasing $B$, $S$, or $L$ can improve the accuracy of the gradient estimate, but can also increase the computational cost simultaneously.

In this section, we leverage the solar data analysis in \cref{sec.solar.flare} as an example to study the trade-off between accuracy and justify the tuning parameter choice in \cref{sec.solar.flare}. 
We experiment with $B \in \{ 5, 10\}$, $L \in \{2, 7\}$ and $S \in \{2, 5, 10\}$, and calculate the log predictive density on held-out data in \cref{tab.tuning}. Overall, \cref{tab.tuning} shows that the predictive performance is insensitive to the choice of our tuning parameters. We record a less than $2\%$ change in log-predictive density as we vary these parameters. 

\begin{table}[ht]
\centering
\begin{tabular}{|c|cccc|cccc|}
\hline
& \multicolumn{4}{c}{$L=2$} & \multicolumn{4}{c}{$L=7$} \vline \\
\hline
&   \multicolumn{2}{c}{$B=5$} & \multicolumn{2}{c}{$B=10$}  &  \multicolumn{2}{c}{$B=5$}  & \multicolumn{2}{c}{$B=10$} \vline \\
\hline
&  Uniform & TASS &  Uniform & TASS &   Uniform & TASS  &  Uniform & TASS  \\
\hline
$S=2$ &  $-9.94$ & $-9.21$ & $-9.86$ & $-9.21$ & $-9.90$ & $-9.22$ & $-9.95$ & $-9.13$ \\
$S=5$ & $-9.89$ & $-9.15$ & $-9.88$ & $-9.06$ & $-9.90$ & $-9.20$ & $-9.92$ & $-9.17$ \\
$S=10$ & $-9.91$ & $-9.18$ & $-9.93$ & $-9.07$ & $-9.87$ & $-9.25$ & $-9.91$ & $-9.16$ \\
\hline
\end{tabular}
\caption{Log predictive density with different tuning parameters (solar flare dataset of \cref{sec.solar.flare}).}
\label{tab.tuning}
\end{table}

In \cref{sec.sleep.cycle}, to determine the tuning parameters, we experiment with $B \in \{ 5, 10\}$, $L \in \{2, 7\}$ and $S \in \{2, 5\}$, and calculate the log predictive density on held-out data. The result is shown in \cref{tab.tuning_sleep}. We record an approximately $10\%$ change in log-predictive density as we vary these parameters.

\begin{table}[ht]
\centering
\begin{tabular}{|c|cccc|cccc|}
\hline
& \multicolumn{4}{c}{$L=2$} & \multicolumn{4}{c}{$L=7$} \vline \\
\hline
&   \multicolumn{2}{c}{$B=5$} & \multicolumn{2}{c}{$B=10$}  &  \multicolumn{2}{c}{$B=5$}  & \multicolumn{2}{c}{$B=10$} \vline \\
\hline
&  Uniform & TASS &  Uniform & TASS &   Uniform & TASS  &  Uniform & TASS  \\
\hline
$S=2$ &  $-2.02$ & $-2.00$ & $-2.01$ & $-1.98$ & $-2.01$ & $-1.75$ & $-2.00$ & $-1.72$ \\
$S=5$ & $-1.94$ & $-1.85$ & $-2.06$ & $-2.00$ & $-2.18$ & $-1.80$ & $-2.05$ & $-1.75$ \\
\hline
\end{tabular}
\caption{Log predictive density with different tuning parameters (sleep cycle data of \cref{sec.sleep.cycle}).}
\label{tab.tuning_sleep}
\end{table}

\end{document}